\documentclass[onecolumn]{robbyant}

\metadata[Website]{\url{https://technology.robbyant.com/lingbot-world}}
\metadata[Github]{\url{https://github.com/robbyant/lingbot-world}}
\metadata[Checkpoints]{\url{https://huggingface.co/robbyant/lingbot-world}}
\newcommand{\methodname}{LingBot-World}
\newcommand{\method}{\texttt{\methodname}\xspace}
\newcommand{\methodbase}{\texttt{\methodname-Base}\xspace}
\newcommand{\methodfast}{\texttt{\methodname-Fast}\xspace}

\newcommand{\E}{\mathbb{E}}

\newcommand{\Loss}{\mathcal{L}}

\title{Advancing Open-source World Models}

\author{
\begin{center}
    {\large Robbyant Team}
\end{center}
}

\begin{document}

\abstract{
We present \method, an \textbf{\textit{open-sourced}} world simulator stemming from video generation. Positioned as a top-tier world model, \method offers the following features. (1) It maintains high fidelity and robust dynamics in a broad spectrum of environments, including realism, scientific contexts, cartoon styles, and beyond. (2) It enables a minute-level horizon while preserving contextual consistency over time, which is also known as ``long-term memory''. (3) It supports real-time interactivity, achieving a latency of under 1 second when producing 16 frames per second. We provide public access to the code and model in an effort to narrow the divide between open-source and closed-source technologies. We believe our release will empower the community with practical applications across areas like content creation, gaming, and robot learning.
}

\maketitle
\begin{figure}[!h]
\centering
\vspace{20pt}
\includegraphics[width=0.98\linewidth]{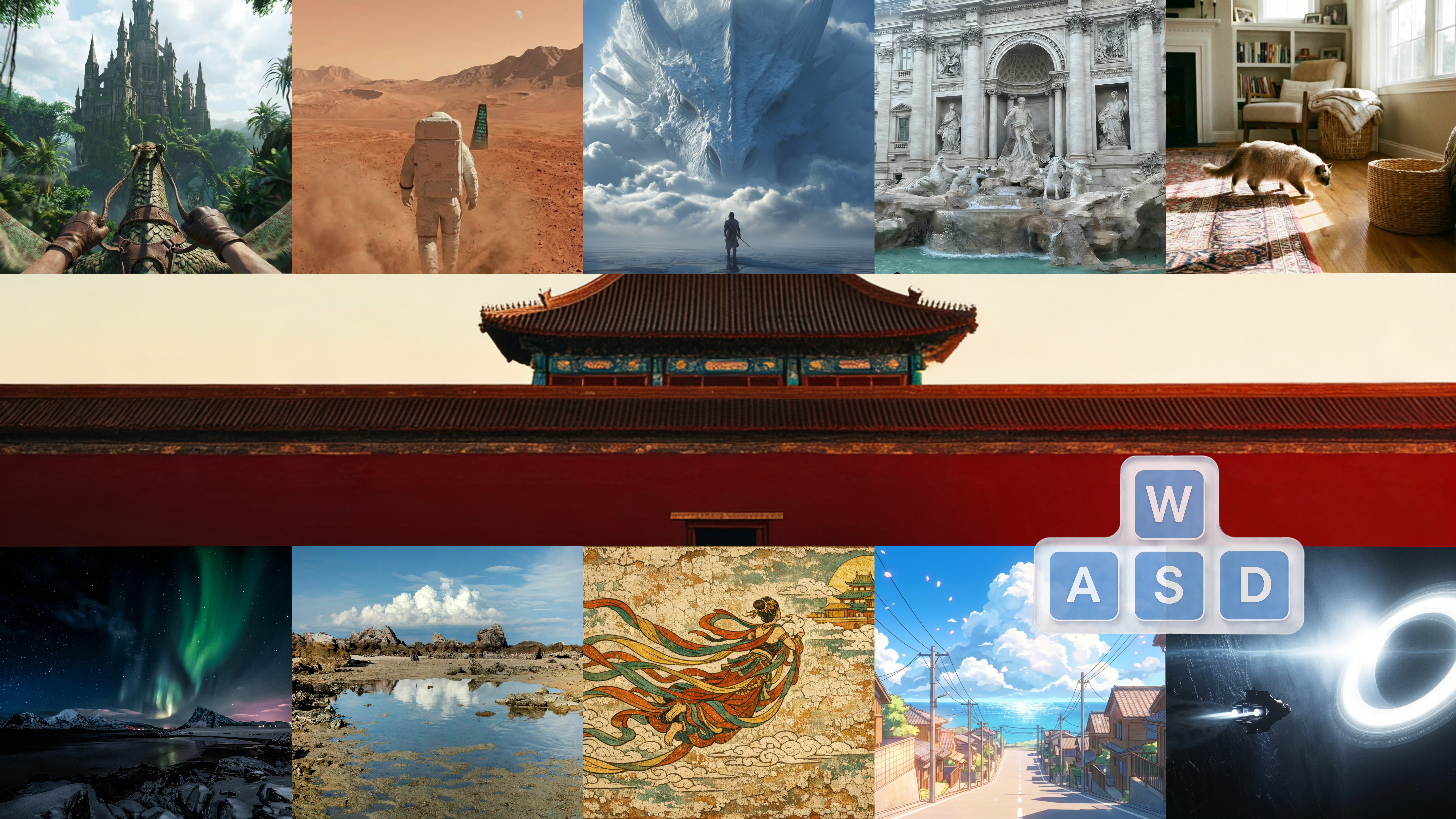}
\caption{\textbf{Interactive world simulation across diverse environments.} The figure showcases selected samples generated by \method, demonstrating its capability to synthesize high-fidelity videos in various domains, including photorealistic landscapes, scientific visualizations, and stylized artistic worlds. The overlaid keyboard icons (\texttt{W}, \texttt{A}, \texttt{S}, \texttt{D}) highlight the model's controllability, allowing users to navigate and interact with these dynamic environments seamlessly.}
\label{fig:teaser}
\end{figure}

\justifying
\section{Introduction}

The pursuit of artificial intelligence capable of understanding and simulating the physical world~\cite{xu2019understanding,long2025survey} has long been considered a ``holy grail'' in computer vision and machine learning. We are currently witnessing a paradigm shift in generative models, transitioning from static ``text-to-video'' generation~\cite{brooks2024video,singer2022make,blattmann2023stable,bar2024lumiere} to the more ambitious goal of ``text-to-world'' simulation~\cite{xiang2025pan,ali2025world,assran2025v,hong2025relic,ha2018world,genie3,mirage2,mao2025yume15,sun2025worldplay,he2025matrix,tang2025hunyuan}. While state-of-the-art video generation models~\cite{wan2025,team2025longcat,kong2024hunyuanvideo,hacohen2024ltx,zhang2025waver,gao2025seedance} have achieved remarkable fidelity in rendering short, visually coherent clips, they fundamentally remain ``dreamers'' rather than ``simulators''~\cite{lecun2022path,brooks2024video}. They hallucinate pixel transitions based on statistical correlations but often lack a grounded understanding of the underlying laws, such as causality, object permanence, and the consequences of interaction. Bridging this gap makes it essential to transition from generating passive footage to building world models capable of synthesizing persistent, interactive, and logically consistent environments.

However, the transition from video generation to world simulation~\cite{valevski2024diffusion,alonso2024diffusion,zhang2025matrix,he2025matrix,spatialvid} faces significant challenges. First, high-quality interactive data are scarce; unlike passive web videos, data that captures the complex interplay between an agent’s decisions and the environment’s reaction is notoriously difficult to scale~\cite{grauman2022ego4d,mialon2023gaia}. Second, maintaining narrative and structural coherence over minute-long trajectories rather than mere seconds—remains an unsolved challenge for standard diffusion architectures, which suffer from ``catastrophic forgetting''~\cite{bar2024lumiere,blattmann2023stable}. Finally, the computational prohibitiveness of traditional diffusion sampling makes live control impossible, limiting most existing models to offline rendering rather than real-time interaction. Furthermore, the most advanced solutions in this space remain proprietary, creating a divide that hinders broader community innovation.

In this report, we present \method, a comprehensive, \textit{open-source} framework designed to shatter these barriers and democratize the research of large-scale world models. \method is not merely a generative model; it is a holistic system engineered to learn the dynamics of virtual worlds and render them in real-time. \method is founded upon three strategic pillars that distinguish our model from existing solutions:

\begin{itemize}
    \item \textbf{A scalable data engine with hierarchical semantics.} We address the data bottleneck by constructing a hybrid engine that ingests diverse data sources, including real-world footage, game engine recordings, and synthetic data from Unreal Engine. Crucially, to solve the lack of fine-grained control in raw data, we introduce a hierarchical captioning strategy~\cite{chen2024panda,chen2024sharegpt4video,wang2025koala}. By generating distinct narrative, scene-static, and dense temporal captions, we effectively disentangle motion control from static scene generation, allowing the model to learn precise action-contingent dynamics.
    \item \textbf{A multi-stage evolutionary training pipeline.} We propose a progressive training strategy to evolve a foundation video generator into an interactive simulator, including three stages: pre-training, middle-training, and post-training. In stage I, a robust general video prior is established via pre-training to support high-fidelity texture generation. In stage II, or middle-training, we employ a mixture-of-experts (MoE) architecture~\cite{damen2018scaling,wan2025,lepikhin2020gshard,fedus2022switch} to incorporate world knowledge and enable action controllability, focusing on ``long-term memory'' and maintaining environmental consistency over extended horizons. In stage III, we optimize the model for real-time inference. Through causal attention adaptation and few-step distillation~\cite{luo2023latent,song2023consistency,salimans2022progressive}, the bidirectional diffusion model is post-trained into an efficient autoregressive system~\cite{bruce2024genie,huang2025self} with sub-second latency.
    \item \textbf{Versatile applications for embodied AI.} Beyond visual synthesis, \method serves as a practical testbed for downstreams~\cite{huang2023voxposer,zitkovich2023rt,hafner2023mastering,feng2025survey,ren2025cosmos,russell2025gaia,valevski2024diffusion,alonso2024diffusion,bar2025navigation}. It supports promptable world events, allowing users to semantically steer global conditions and local dynamics via textual prompts. Furthermore, it facilitates the training of action agents and enables consistent 3D reconstruction from generated videos~\cite{kerbl20233d,mildenhall2021nerf,xie2024physgaussian}, validating its geometric integrity.
\end{itemize}

To contextualize our contribution,~\cref{tab:comparison} compares \method with recent interactive world models. While systems like Genie 3~\cite{genie3} and Mirage 2~\cite{mirage2} have made strides, they often compromise on dynamic degree or remain closed-source. \method distinguishes itself by offering a \textbf{\textit{general domain}} capability, a \textbf{\textit{long generation horizon}}, and a \textbf{\textit{high dynamic degree}} in \textbf{\textit{real-time}}, all while being fully \textbf{\textit{open-source}}. By releasing the code and model weights, we aim to ignite a new wave of innovation, empowering the community to build the next generation of virtual worlds.

\begin{table}[t]
    \small
    \centering
    \caption{
        \textbf{Comparison with recent interactive world models.} \method stands out by combining high dynamic degree and long generation horizons within a general domain, while being the only high-capability model that is fully open-sourced.
    }
    \label{tab:comparison}
    \SetTblrInner{rowsep=1.2pt}      
    \SetTblrInner{colsep=4.6pt}     
    \definecolor{linegray}{HTML}{BDBDBD} 
    \definecolor{bg_gray1}{HTML}{FAFAFA}
    \definecolor{bg_gray2}{HTML}{F2F2F2}
    \definecolor{bg_gray3}{HTML}{EAEAEA}
    \definecolor{bg_gray4}{HTML}{E2E2E2}
    \definecolor{bg_gray5}{HTML}{DADADA}
    \definecolor{bg_purple}{HTML}{6A67F3}
    \begin{tblr}{
        cells={halign=l,valign=m},
        column{1}={bg=white},
        column{2}={bg=bg_gray1},
        column{3}={bg=bg_gray2},
        column{4}={bg=bg_gray3},
        column{5}={bg=bg_gray4},
        column{6}={bg=bg_gray5},
        column{7}={bg=bg_purple, fg=white},
        hline{2}={0.5pt, fg=linegray},
    }
    \ & \textbf{Matrix-Game 2.0}~\cite{he2025matrix} & \textbf{Yume-1.5}~\cite{mao2025yume15} & \textbf{HY-World 1.5}~\cite{sun2025worldplay} & \textbf{Mirage 2}~\cite{mirage2} & \textbf{Genie 3}~\cite{genie3} & \textbf{Ours} \\
    Domain & Game & General & General & General & General & General \\ 
    Generation Horizon & Short & Short & Medium & Long & Long & Long \\ 
    Dynamic Degree & Low & Low & Low & Medium & Medium & High \\ 
    Resolution & 480p & 480p & 720p & 480p & 720p & 720p \\ 
    Real-time & \ding{51} & \ding{55} & \ding{51} & \ding{51} & \ding{51} & \ding{51} \\ 
    Open-source & \ding{51} & \ding{51} & \ding{51} & \ding{55} & \ding{55} & \ding{51} \\ 
    \end{tblr}
\end{table}

By open-sourcing \method, including our model weights and inference codebase, we aim to ignite a new wave of innovation. We invite the community to move beyond passive video watching and join us in building the next generation of infinite, playable, and interactive virtual worlds.
\section{Data Engine} \label{sec:data}

Constructing a world model capable of robustly handling novel viewpoints, complex dynamics, and long-horizon planning requires a rigorous data strategy. We address this by structuring our data engine as a unified pipeline with three synergistic components: (i) \textbf{data acquisition}, (ii) \textbf{data profiling}, and (iii) \textbf{data captioning}.

\begin{figure}[t]
\centering
\includegraphics[width=0.85\linewidth]{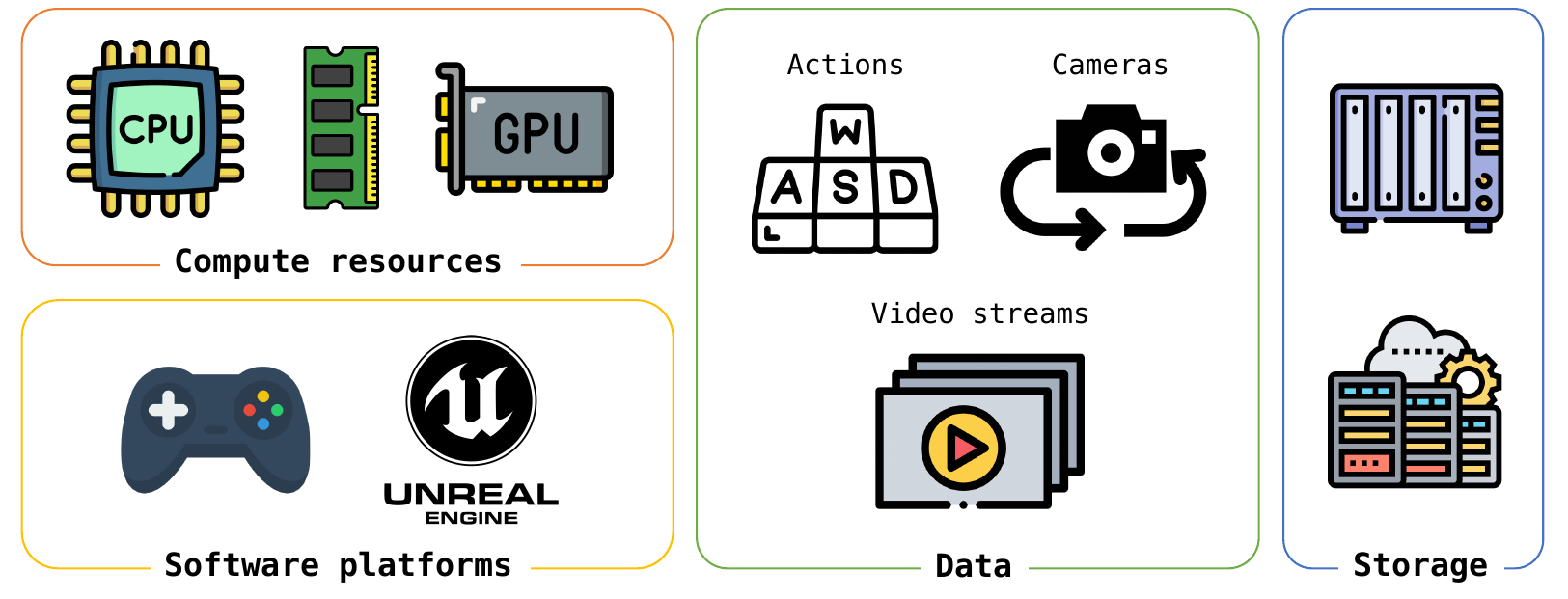}
\caption{\textbf{Game and synthetic data acquisition.} The system leverages computational resources and software platforms to capture visual observations that are temporally aligned with action signals and camera states.}
\label{fig:data}
\end{figure}

To build the foundation of this system, our \textbf{data acquisition} phase employs a hybrid collection strategy designed to guarantee a rich, high-quality, and interactive training corpus. First, we curate a large-scale dataset of high-quality, diverse videos, featuring both first-person~\cite{grauman2022ego4d,damen2018scaling} and third-person~\cite{soomro2012ucf101,bain2021frozen} perspectives of humans, animals, and vehicles. Second, to capture precise action-contingent dynamics, we harvest game data where RGB frames are strictly paired with user control inputs (e.g., \texttt{W}, \texttt{A}, \texttt{S}, \texttt{D}) and camera parameters. Finally, we develop a synthetic rendering pipeline using Unreal Engine (UE)~\cite{ue}. By integrating licensed assets with custom-built environments, we design an automated rendering workflow that generates collision-free, randomized yet plausible camera trajectories, yielding RGB streams aligned with ground-truth camera intrinsics and extrinsics. The high-level idea is depicted in~\cref{fig:data}.

After acquisition, the \textbf{data profiling} component acts as a critical standardization layer. To unify the diverse inputs, where general videos lack the camera information compared to game or UE data, we first utilize state-of-the-art pose estimation models~\cite{schonberger2016structure,kendall2015posenet,sarlin2020superglue} to generate pseudo-labels for camera intrinsics and extrinsics. Then, the system executes basic filtering to discard substandard samples based on resolution and duration, while employing off-the-shelf algorithms to slice footage into training-friendly clips~\cite{pyscenedetect,transnetv2}. Finally, we utilize a vision-language model (VLM)~\cite{radford2021learning,achiam2023gpt,team2023gemini} to perform semantic analysis, evaluating attributes such as visual quality, motion magnitude, and scene perspective to curate the filtered dataset.

Following acquisition and filtering, the \textbf{data captioning} module finally enriches the corpus with semantic metadata using a vision-language model (VLM)~\cite{bai2025qwen3,lu2024deepseek}. We implement a hierarchical annotation strategy that produces three distinct layers of description to ensure a multi-granular understanding of the video content. This includes a \textit{comprehensive narrative} caption that weaves environment and camera movement into a global story, a \textit{scene-static} caption that focuses purely on the environment, and \textit{dense temporal} captions that offer fine-grained, time-aligned accounts of specific events.

\subsection{Data Acquisition}

\subsubsection{General Video Curator}
Given the vast quantity of raw video data available from both in-house collections and open-source repositories, effective data selection is critical~\cite{peebles2023scalable,brooks2024video,gupta2024photorealistic}. We develop a \textbf{general video curator} designed to filter and retrieve high-value samples that align with our specific training objectives. This curation process prioritizes the video content category of \textit{diverse world exploration}, aiming to maximize the breadth of motion patterns and environmental contexts. This includes a wide spectrum of locomotion types (e.g., walking, cycling, transit) captured from diverse viewpoints, ranging from human and animal ego-centric perspectives to third-person camera angles.

\subsubsection{Game Data Acquisition}
\label{sec:game_data}
We develop a dedicated \textbf{game data acquisition platform} engineered for high-fidelity capture and synchronization of visual data, agent actions, and camera movement~\cite{dxc}. To ensure a pristine visual baseline, the display environment is configured to exclude interface overlays, ensuring consistent visual quality via appropriate codecs. User control signals are registered with high-precision timestamps to guarantee synchronization with video frames. Furthermore, the designed camera trajectories are recorded to ensure reliable geometric information.

To ensure our game data covers a diverse range of behaviors and environmental complexities, we establish a standardized collection strategy divided into four primary categories:
\begin{itemize}
    \item \textbf{Navigation:} Covers general movement through the virtual world. 
    \begin{itemize}
        \item \textit{Free navigation}: Enabling stochastic exploration across random trajectories; 
        \item \textit{Loop roaming}: Recording closed-loop paths or multi-point round trips; 
        \item \textit{Transition navigation}: targeting high-variance scene changes, such as exiting buildings or switching between distinct interior environments.
    \end{itemize}
    
    \item \textbf{Sightseeing:} Focuses on fine-grained observation. This involves carefully examining scene details in both static and dynamic environments, as well as orbiting around landmark objects to capture multi-view consistency.
    
    \item \textbf{Long-tail scenarios:} Targets rare but critical data distributions often missing from standard ones.
    \begin{itemize}
        \item \textit{Stationary observation:} Capturing data from a fixed position without translational movement, including 360-degree rotation to map static surroundings and fixed-angle staring to record dynamic elements (e.g., crowds or traffic) evolving over time.
        \item \textit{Backward navigation:} Retreating while maintaining situational awareness.
    \end{itemize}
    
    \item \textbf{World interaction:} Captures causal agent-environment relationships, ranging from localized actions (e.g., picking up items, opening doors) to impactful events triggering significant state changes (e.g., combat, destruction).
\end{itemize}

\subsubsection{Synthetic Rendering Pipeline}
Our synthetic rendering pipeline utilizes Unreal Engine~\cite{ue} to generate scalable synthetic datasets with precise camera poses and customizable navigation trajectories. This pipeline offers two primary advantages: First, it provides accurate and temporally aligned camera poses required for effective action training—a level of fidelity often unattainable with real-world sensors. Second, it allows us to expand trajectory diversity beyond the biases of real-world datasets, which are frequently dominated by simplistic patterns such as forward motions, and augment the density of re-observation trajectories required for spatial memory.

To realize these capabilities, we develop a streamlined automated workflow. The process begins by randomly sampling a semantically meaningful position and orientation within a scene to serve as a start point. From this location, the system automatically generates a camera trajectory by either sampling from randomized parameters or leveraging imported motion priors. Each generated trajectory undergoes rigorous collision detection. Finally, the confirmed trajectory is processed for video rendering alongside the export of synchronized ground truth camera poses.

To populate this workflow with varied and realistic motions, the trajectory generation stage operates in two distinct modes designed to balance stochastic diversity with behavioral authenticity. 

\begin{itemize}
    \item \textbf{Procedural path generation:} This mode autonomously synthesizes complex camera movements to maximize environmental exploration, focusing on two primary algorithmic strategies:
    \begin{itemize}
        \item \textit{Geometric pattern synthesis:} The system generates structured trajectories, including randomized rectangular paths of varying scales and multi-turn 360$^\circ$ rotations at diverse angular velocities. These patterns provide comprehensive panoramic context and reinforce long-term spatial consistency through repetitive environmental coverage.
        \item \textit{Multi-point interpolation:} This strategy samples random spatial waypoints with reciprocal look-back transitions, which specifically strengthens relational spatial memory.
    \end{itemize}
    
    \item \textbf{Real-world trajectory import:} This mode maps paths captured from physical devices directly into Unreal Engine. It incorporates authentic human browsing behaviors, such as repeatedly scanning a room or revisiting objects of interest, while retaining the subtle nuances of handheld motion (e.g., natural jitter and organic velocity changes) to reflect the stochasticity and temporal complexity characteristic of actual user interaction.
\end{itemize}

\subsection{Data Profiling}

\begin{figure}[t]
\centering
\includegraphics[width=\linewidth]{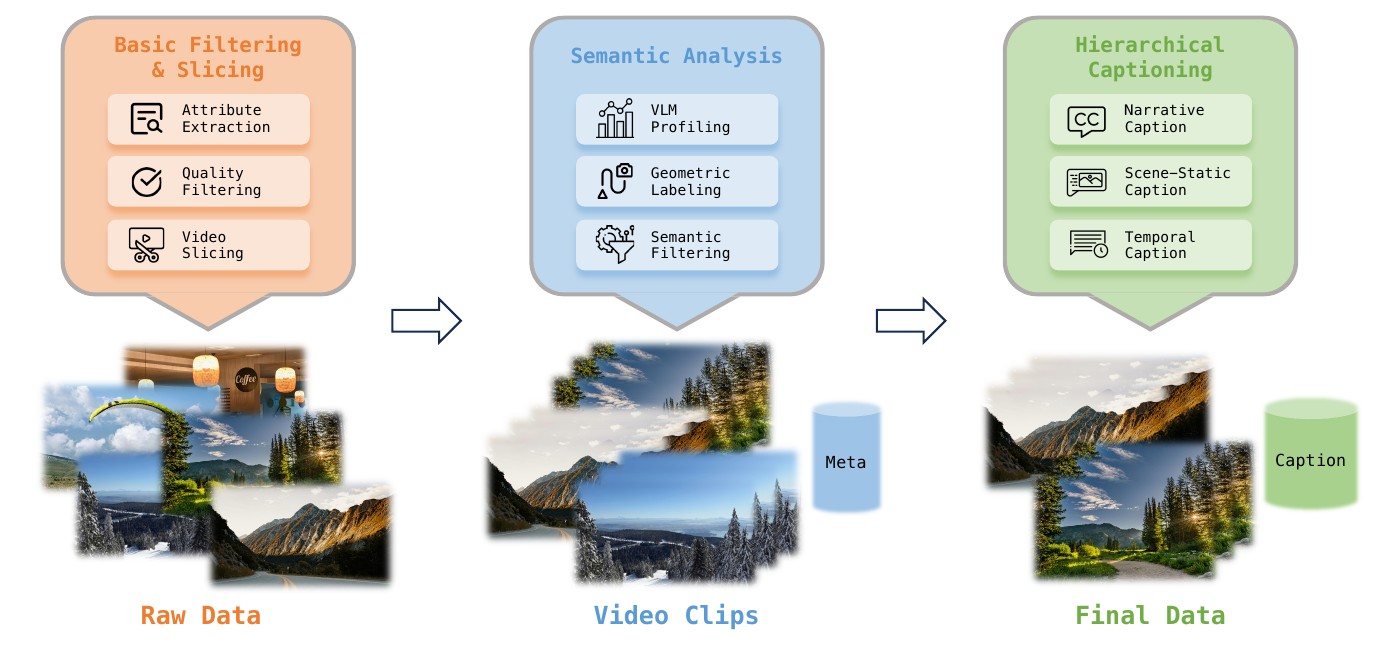}
\caption{\textbf{Overview of data profiling engine.} The process bridges the gap between raw video collections and training-ready assets. It integrates physical attribute filtering, semantic profiling, and geometric annotation to establish a robust foundation for the subsequent hierarchical captioning generation.
}
\label{fig:datapipeline}
\end{figure}

Following data acquisition, the \textbf{data profiling engine} performs a comprehensive analysis to extract multi-dimensional metadata for each video. This process operates on three distinct levels of granularity, as illustrated in~\cref{fig:datapipeline}.

\subsubsection{Basic Filtering \& Temporal Segment} 
At the fundamental level, we extract \textit{basic file attributes}, including video duration, resolution, and file size, to establish a fundamental index of the dataset. Guided by this metadata, we eliminate substandard samples, specifically those with insufficient resolution or inadequate duration. Subsequently, we utilize the slicing algorithm provided by Koala~\cite{wang2025koala} alongside TransNet v2~\cite{transnetv2} to segment the footage into training-friendly clips. This approach preserves the semantic coherence and consistency of each segment, ensuring a high-quality video source for downstream processing.

\subsubsection{Semantic Analysis}
Advancing to semantic analysis, we employ our internal vision-language model (VLM) to extract a comprehensive set of \textit{filtering attributes}. Specifically, the model evaluates visual quality (including brightness and sharpness), quantifies motion magnitude, and identifies scene types and perspectives (e.g., first-person vs. third-person). These semantic descriptors provide a robust basis for precise data selection in downstream processing.

To address the lack of geometric information in raw videos, we further utilize MegaSAM~\cite{megasam} to generate \textit{camera pose annotations} for videos lacking geometric information. This ensures that all selected samples possess the necessary 3D structural priors required for training.

Ultimately, this two-stage profiling strategy bridges the gap between raw video collections and training-ready assets. By layering basic physical attributes with high-level semantic descriptors and intrinsic geometric data, we establish a robust foundation for the subsequent training phases.

\subsection{Data Captioning}
With the data curated and profiled, we introduce the \textbf{hierarchical captioning module}. To effectively train world models, we design a specialized annotation strategy that goes beyond simple tagging. We generate three distinct categories of captions for each video, catering to different granularities of semantic control and motion decoupling:

\begin{enumerate}
    \item \textbf{Comprehensive narrative caption:} This type provides a holistic and detailed description of the entire video, intertwining the visual environment with the camera's trajectory and temporal evolution. It serves as a global semantic prompt.
    
    \begin{quote}
    \small
    \textit{\textbf{Example:} The video unfolds as a tranquil, first-person exploration of a meticulously designed East Asian-style courtyard or temple interior. The journey begins by approaching a set of richly painted wooden screens depicting phoenixes, hinting at the cultural significance of the space. As the camera glides forward and pans left, it reveals the depth of the interior, showcasing a towering striped column, softly glowing lanterns, and a majestic white statue resting on an ornate pedestal, all bathed in warm, ambient light. The perspective then shifts rightward, guiding the viewer along a colonnaded walkway with textured stone walls toward imposing red doors studded with gold, which serve as both a focal point and a potential threshold to the outside world. Continuing deeper, the camera navigates a quieter side corridor where lantern-lit windows cast gentle illumination on the cracked stone floor, enhancing the sense of aged serenity. A deliberate turn brings the viewer back to admire the central statue once more, its presence emphasized by the dramatic play of light and shadow on the ground. Finally, the camera retraces its path, returning to the grand doors and then back to the initial screens, completing a circular tour that invites contemplation of the architecture’s symmetry, detail, and peaceful atmosphere—all captured through smooth, unhurried movements that emphasize immersion and visual discovery.}
    \end{quote}

    \item \textbf{Scene-static caption (action-decoupled):} This caption focuses exclusively on the static environment and aesthetic details, deliberately omitting descriptions of camera movement or character actions. This design is crucial for decoupling motion control from scene generation in world models.
    
    \begin{quote}
    \small
    \textit{\textbf{Example:} The video presents a first-person perspective of someone wandering through a serene, ornately decorated courtyard or temple complex with traditional East Asian architectural elements. The environment features textured stone walls, intricately painted wooden screens, large red doors with golden studs, and a central statue on a pedestal, all under soft, ambient lighting that casts long shadows across the cracked stone pavement. The atmosphere is calm and still, with no other characters or moving objects present, emphasizing the quiet beauty and detailed craftsmanship of the surroundings.}
    \end{quote}

    \item \textbf{Dense temporal caption:} This type offers fine-grained, time-aligned descriptions by segmenting the video into intervals and detailing events within each to support temporal alignment training.
    
    \begin{quote}
    \scriptsize
    \begin{verbatim}
[
  {
    "start_time": 0.0, "end_time": 5.0, 
    "Event": "Approaching decorative screen", 
    "caption": "The camera moves forward toward a set of ornate wooden 
    screens featuring painted phoenixes, positioned at the entrance to 
    a raised area with steps. To the left, stacks of green and red 
    cylindrical objects are visible inside the structure."
  }, 
  {
    "start_time": 5.0, "end_time": 10.0, 
    "Event": "Panning left to reveal interior", 
    "caption": "The camera pans left, revealing more of the interior 
    space, including a tall, striped column, hanging lanterns, and a 
    glimpse of a large white statue on a decorated pedestal in the 
    background."
  }, 
  {
    "start_time": 10.0, "end_time": 15.0, 
    "Event": "Moving toward large doors", 
    "caption": "The camera turns right and moves along a corridor with 
    textured stone walls and wooden pillars, approaching a pair of 
    large, imposing red doors adorned with golden circular patterns 
    and black metal studs."
  }, 
  ...
  {
    "start_time": 30.0, "end_time": 35.0, 
    "Event": "Revisiting decorative screen", 
    "caption": "The camera returns to the initial position facing the 
    ornate wooden screens with phoenix paintings, providing a 
    symmetrical bookend to the exploration loop."
  }
]
    \end{verbatim}
    \end{quote}
\end{enumerate}

Collectively, this hierarchical captioning framework ensures that every video clip is paired with a rich, structured textual condition. These annotations not only capture static visual details but also encode dynamic evolution and camera intent.
\section{LingBot-World}\label{sec:method}

\subsection{Formulation}
We formulate the world model as a conditional generative process that simulates the evolution of visual states driven by agent actions. Let $\mathcal{V} = \{x_1, x_2, \dots, x_T\}$ denote a sequence of video frames, where $x_t \in \mathbb{R}^{H \times W \times C}$ represents the state at time step $t$. Let $\mathcal{A} = \{a_1, a_2, \dots, a_{T}\}$ denote the corresponding sequence of control signals (actions).

The goal of \method is to learn a parametric model $\theta$ that approximates the transition dynamics of the environment. Instead of strictly limiting the model to single-step prediction, we formulate the objective generally as maximizing the likelihood of future states given the history frames and the current control signals:
\begin{equation}
\label{equ:formulation}
    \max_\theta \mathbb{E} \left[ \log p_\theta(x_{t:t+L} \mid x_{<t}, a_{t:t+L}) \right],
\end{equation}
where $L \ge 1$ represents the prediction horizon. To bridge the gap between a standard video generator and an efficient, interactive world simulator, we propose a \textbf{multi-stage evolution strategy}, decomposing the learning process into three progressive stages: foundation, knowledge injection, and interaction readiness.

\begin{figure*}[t]
\centering
\includegraphics[width=\linewidth]{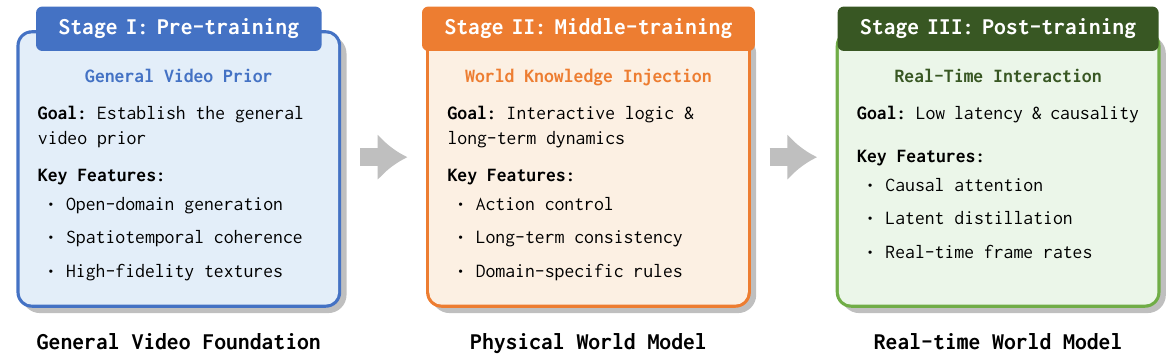}
\caption{\textbf{Overview of the LingBot-World training pipeline.} 
We propose a multi-stage evolution strategy to transform a foundation video generator into an interactive world simulator. 
\textbf{Pre-training} stage establishes a robust general video prior to ensure high-fidelity open-domain generation. 
\textbf{Middle-training} stage injects world knowledge and action controllability, enabling the model to simulate long-term dynamics with consistent interactive logic. 
\textbf{Post-training} stage adapts the architecture for real-time interaction, employing causal attention and few-step distillation to achieve low latency and strict causality.}
\label{fig:general_pipeline}
\end{figure*}

\begin{figure*}[t]
\centering
\includegraphics[width=\linewidth]{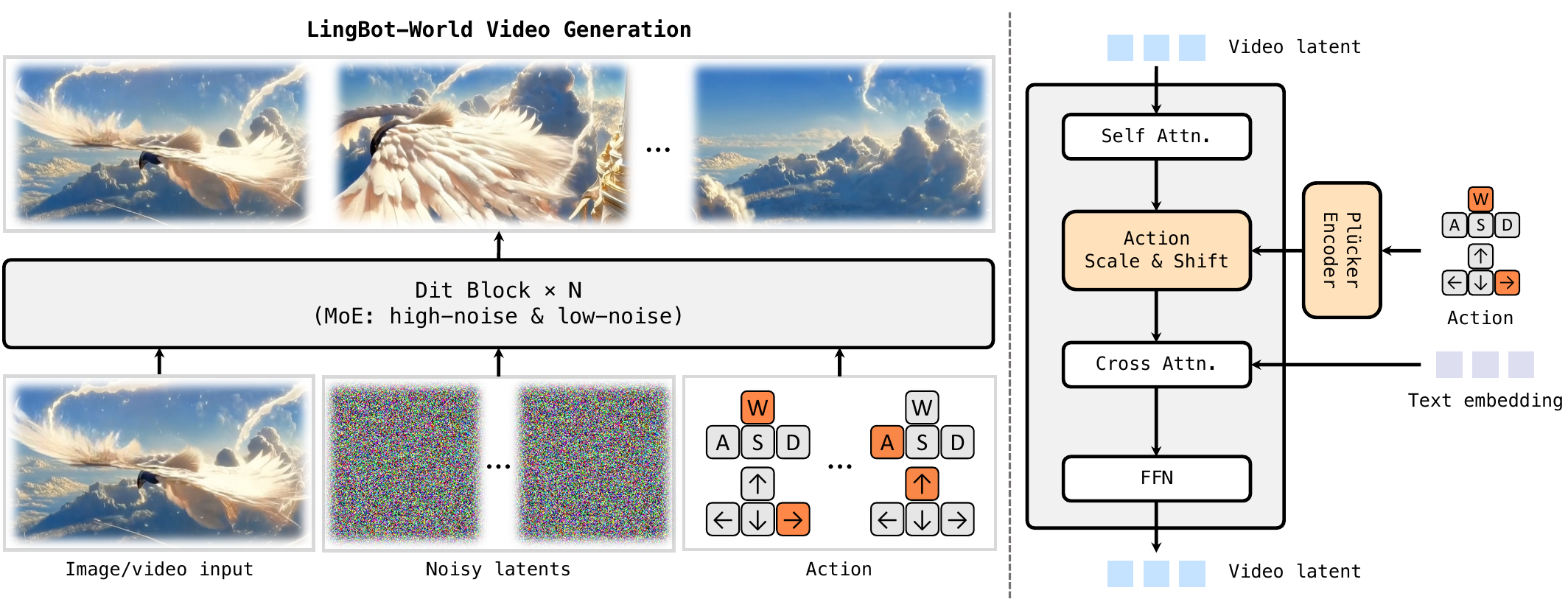}
\caption{\textbf{Pipeline of LingBot-World.}
The left part shows the pipeline of \method video generation.
\method uses an image or a video, noisy latents, and user-defined action signals as inputs to generate video sequences with spatial memory, long-term consistency, and precise action following.
The right part shows the architecture of the DiT blocks in \method.
The video latent first passes through a self-attention layer, enabling \method to learn spatiotemporal coherence, and further emerge spatial memory ability.
Then, action signals are injected through a Plücker Encoder, where the input actions are projected into Plücker embeddings and modulate the video latent via adaptive normalization that transforms the Plücker embeddings into scaling and shifting factors.
Finally, a cross-attention layer is applied to condition the video latent on text embeddings.
}
\label{fig:mid_pipeline}
\end{figure*}

\subsubsection{Stage I: Pre-Training --- Establishing the General Video Prior}
In the initial stage, we focus on modeling the unconditional distribution of natural video sequences and establishing the general prior over visual dynamics.
To this end, we leverage a base video generator pretrained on large-scale open-domain video data, which endows \method with strong \textbf{spatiotemporal coherence} and \textbf{open-domain semantic understanding}.
This pretrained model enables the synthesis of \textbf{high-fidelity object textures} and coherent scene structures, serving as a general visual ``canvas'' for subsequent interactive dynamics rather than encoding task-specific physical rules.

\subsubsection{Stage II: Middle-Training --- Injecting World Knowledge \& Long-Term Dynamics}
In the middle stage, we lift our initial generator into a \textbf{bidirectional world model}.
Under our general formulation~\cref{equ:formulation}, setting $t=0$ aligns with the bidirectional paradigm, allowing the model to first capture global temporal dependencies without being constrained by causal history.
While the pretrained model has shown great potential in high-fidelity video generation, it is limited to short clips and lacks interactive logic. 
Therefore, we further train \method with a specialized data engine to incorporate \textbf{action control}, \textbf{temporal consistency}, and \textbf{domain-specific rules}. The key improvements introduced in this stage are as follows:
\begin{itemize}
    \item \textbf{Long-term consistency:} 
    To enhance the memory capacity, the model is trained on extended video sequences. 
    By observing long-term contextual frames, \method learns to mitigate the “forgetting” problem during video generation, ensuring that the generated visual world remains coherent over minutes of gameplay rather than just seconds.
    \item \textbf{Action controllability:} 
    To introduce interactive capacity, we incorporate user-defined action signals into the model through adaptive normalization~\cite{xu2019understanding, wan2025}. 
    Conditioned on these explicit action inputs, \method generates a visual world that is no longer driven by stochastic noise but follows user-specified instructions.
    \textit{Remark:} At this stage, the model operates as a \textbf{holistic world simulator}, capable of generating high-fidelity future trajectories conditioned on actions, though it relies on bidirectional attention, which is computationally heavy for real-time rollout.
\end{itemize}

\subsubsection{Stage III: Post-Training --- Causal Architecture Adaptation \& Few-Step Distillation}
The final stage transforms our bidirectional world model into an efficient autoregressive system capable of real-time interactive generation. 
By generalizing~\cref{equ:formulation} to $t \ge 0$ and conditioning on past context $x_{<t}$, our formulation seamlessly shifts to the causal paradigm, enabling the step-by-step inference required for interaction. While the Stage II model captures the world dynamics accurately, standard bidirectional diffusion models are computationally prohibitive for deployment due to full temporal attention and multi-step iterative denoising. We address these limitations through:

\begin{itemize}
    \item \textbf{Causal architecture adaptation:} We replace full temporal attention with block causal attention, combining local bidirectional dependencies within chunks and global causality across chunks. The model, initialized from the high-noise expert (Stage II), is trained with a mixed-timestep protocol to bridge expert specialization. This enables efficient autoregressive generation via KV caching while preserving temporal coherence.
    \item \textbf{Few-step distillation:} 
    We employ distribution matching distillation (DMD) augmented with self-rollout training and adversarial optimization \cite{yin2024improved,yin2024one,apt1}. This dual approach distills a few-step generator that maintains action-conditioned dynamics and visual fidelity across extended rollouts without significant drift.
\end{itemize}

\subsection{Pre-Training}
\label{sec:pretraining}
The goal of the pre-training stage is to find a pre-trained model and provides strong video prior for subsequent stages, enabling \method to generate diverse, coherent, and high-fidelity videos. 
Recent advances in world models~\cite{hafner2023mastering,bar2025navigation,chen2025vl}, such as Genie 3~\cite{genie3}, have demonstrated the effectiveness of initializing the interactive world model from powerful video foundation models. 
These video foundation models~\cite{agarwal2025cosmos,tang2025hunyuan,chen2025seedance,hacohen2024ltx,teng2025magi} can provide strong internal priors (i.e., visual realism, object permanence, and temporal dynamics) that can significantly accelerate the learning of interactive physics and controllable visual world generation. 
To this end, we adopt the 14B-parameter Wan2.2 image-to-video diffusion model~\cite{wan2025} as our pre-trained model, which is particularly well-suited for capturing complex spatiotemporal consistency and generating high-fidelity video content.

\subsection{Middle-Training}
In the middle-training stage, the pretrained video diffusion model is transformed into a bidirectional world model to generate a coherent and interactive visual world. 
While the pretrained model demonstrates strong performance in high-fidelity video synthesis, it is inherently limited to short video clips and lacks the ability to interact with user-defined action signals.
To address these limitations, we leverage the proposed data engine (\cref{sec:data}) to generate action-conditioned, temporally extended video sequences for the middle-training stage. 
This stage consists of three primary components.
First, a fundamental world model is trained to acquire long-term temporal consistency and emergent spatial memory, ensuring the stability of the generated world~(\cref{sec:base_model}).
Second, we finetune this fundamental world model to support action-conditioned generation by injecting user-defined action signals into the DiT blocks, enabling controllable dynamics (\cref{sec:action_model}).
Third, as training the fundamental world model is computationally intensive, we implement a parallelism infrastructure that enables efficient training while keeping GPU memory consumption within practical limits (\cref{sec:infra}).
Through this middle-training stage, \method gradually learns long-term temporal consistency, spatial memory, and precise action-conditioned dynamics, bridging the gap between random video generation and interactive, controllable world modeling.

\subsubsection{Fundamental World Model}
\label{sec:base_model}
As shown in~\cref{fig:mid_pipeline}, \method takes an image or a video, noisy latents, and user-defined actions as inputs to generate a controllable visual world instead of random video synthesis.
We first train a fundamental world model that, given an arbitrary initial state (i.e., a single image or a video), generates a visual world exhibiting both long-term video consistency and spatial memory. The training strategies are as follows:

\noindent\textbf{Mixture-of-experts (MoE) architecture.} 
Following the Wan2.2 image-to-video diffusion model~\cite{wan2025}, which has demonstrated the effectiveness of MoE~\cite{mu2025comprehensive} architecture, \method inherits its MoE design to improve model performance while keeping inference cost nearly unchanged.
Since different denoising stages serve their own roles, \method adopts a two-expert design tailored to the diffusion process: a high-noise expert, activated at early timesteps, focuses on modeling global structure and coarse layout, while a low-noise expert, activated at later timesteps, polishes fine-grained spatial and temporal details.
Each expert contains approximately 14B parameters, resulting in a total model size of 28B parameters, while only one expert is activated at each denoising timestep. This design preserves inference-time computation and GPU memory consumption comparable to a dense 14B model.

\noindent\textbf{Progressive curriculum training.} 
To enable \method to achieve long-term video consistency and spatial memory, we adopt a progressive curriculum training strategy.
In the first round, we employ 5-second video sequences to train the fundamental world model and broaden its internal generation domain, as the pretrained model is constrained by a narrow distribution.
Then, we progressively extend the training duration from 5 seconds to 60 seconds, allowing the fundamental world model to learn long-term temporal consistency and facilitate the emergence of spatial memory.
Furthermore, we progressively scale the flow shift in correspondence with the increasing video duration.
This design is motivated by the observation that long-video generation requires a greater emphasis on high-noise timesteps, which are responsible for modeling global scene structure. 
Increasing the proportion of high-noise timesteps helps stabilize scene-level structure over extended temporal ranges, thus reducing drift and improving performance in long-term video generation.

\noindent\textbf{Multi-task training.} 
To endow \method with the ability to predict future world states from arbitrary initial conditions, we adopt a multi-task training paradigm incorporating both image-to-video and video-to-video (i.e., video continuation) objectives. 
These tasks correspond to different forms of initial states: the image-to-video task enables the fundamental world model to infer future dynamics from a single static frame, while the video-to-video task facilitates extrapolation beyond observed motion by predicting future frames from historical sequences. 
By jointly optimizing these complementary tasks, the model learns a unified world transition function that generalizes across diverse initial conditions, allowing robust prediction of future world states from arbitrary starting points in time.

\subsubsection{Action-Conditioned World Model}
\label{sec:action_model}

After training the fundamental world model to establish long-term temporal consistency and spatial memory, we proceed to finetune the model to support interactive control. This stage transforms the video generator into a responsive world simulator by injecting user-defined action signals.

\noindent\textbf{Action representation.}
To enable precise control over the generated environment, we employ a hybrid action representation strategy that combines continuous camera rotation with discrete keyboard inputs (e.g., \texttt{W}, \texttt{A}, \texttt{S}, \texttt{D}). Specifically, we represent camera rotation using Plücker embeddings, which provide a geometric representation suitable for continuous 3D transformations. Simultaneously, discrete interactions are encoded as multi-hot vectors. These two modalities are fused via concatenation along the channel dimension. This hybrid representation ensures the model can handle both smooth view changes and distinct logical state transitions.

\noindent\textbf{Action injection mechanism.}
To incorporate these action signals into the diffusion process without disrupting the pre-trained visual priors, we utilize an adaptive layer normalization (AdaLN) mechanism~\cite{xu2019understanding}. The fused action embeddings are projected and injected into the DiT blocks. This allows the action signals to modulate the normalized features dynamically, guiding the denoising process to generate video frames that are consistent with the specified actions.

\noindent\textbf{Finetuning paradigm.}
We adopt a parameter-efficient finetuning strategy to preserve the generative quality of the fundamental model. Specifically, we freeze the main DiT blocks of the pre-trained fundamental world model and only finetune the newly added action adapter layers (including the action embedding projections and AdaLN parameters).
This design is driven by two key motivations: (1) It effectively disentangles the inherent video generation capability from the action control capability. (2) Since high-quality action-labeled data is often rare or synthetic (generated via the data engine mentioned in~\cref{sec:game_data}), fully finetuning the dense model carries a risk of catastrophic forgetting or degradation of the fundamental visual quality. Freezing the backbone ensures the model retains its high-fidelity video synthesis abilities while learning to follow control signals.

\subsubsection{Parallelism Infrastructure}
\label{sec:infra}
Training \method, a 28B-parameter fundamental world model, on one-minute video sequences is highly demanding in terms of GPU memory. 
This is due to the combination of the large model size, long token length, and memory-intensive operations such as gradient computation, optimizer state management, and activation checkpointing. 
To overcome these challenges, we implement a parallelism infrastructure that efficiently distributes computation and memory across multiple GPUs.

\noindent\textbf{Fully sharded data parallel 2 (FSDP2).} 
To support efficient training of 28B-parameter \method, we employ FSDP2~\cite{zhao2023pytorch} for scalable data parallelism. 
FSDP2 employs a fully sharded scheme where each GPU holds only a fraction of the model parameters, gradients, and optimizer states, enabling the training of large-scale models that would otherwise exceed single-GPU memory limits.
Moreover, by overlapping communication with computation and leveraging other system-level optimizations, FSDP2 achieves high training efficiency and near-linear throughput scaling as both model size and GPU counts increase.

\noindent\textbf{Context parallel (CP).} 
To mitigate the memory bottleneck arising from long token length, we adopt Ulysses~\cite{jacobs2023deepspeed} for context parallel strategy. 
Ulysses introduces sequence parallelism by partitioning the input tensor along the temporal~(sequence) dimension and distributing these slices across multiple GPUs. 
During attention computation, an efficient all-to-all collective communication pattern redistributes the necessary activations such that each device can locally compute attention over its sequence shard. 
By sharding the sequence dimension in this way, the per-GPU memory footprint for activations and attention-related intermediates is significantly reduced, allowing \method to process long sequences in parallel.

\subsection{Post-Training}
In the post-training stage, we transform our bidirectional world model into an efficient autoregressive model capable of real-time interactive generation. This transformation addresses the computational constraints of deploying bidirectional attention for real-time applications while preserving the rich dynamics learned during middle-training. 
Our post-training methodology consists of two key stages. First, we adapt the bidirectional architecture into a causal framework through diffusion forcing mechanism (\cref{subsec:ode}) \cite{chen2024diffusion}. Second, we employ few-step distillation augmented with long-horizon training to transfer the teacher's capabilities to the student model (\cref{subsec:dmd}) \cite{huang2025self,apt1}. 
Throughout this process, we prioritize preserving two critical competencies. We maintain accurate action-conditioned dynamics modeling and ensure sustained visual fidelity across extended temporal sequences without accumulative drift.

\subsubsection{Causal Architecture Adaptation} 
\label{subsec:ode}

\noindent\textbf{Model initialization.} 
Recall that our middle-trained model is a mixture-of-experts image-to-video diffusion model comprising two sequential experts: a high-noise expert and a low-noise expert. Each expert specializes in denoising specific timestep ranges of the diffusion process. For simplified and efficient training and inference, we initialize our causal student model using the high-noise expert due to its superior dynamics modeling capabilities.
The initialization from our middle-trained model provides inherent advantages through progressive curriculum learning. The model already possesses the ability to attend to variable-length token sequences, which makes our causal adaptation more stable and generalizable to rollouts of varying lengths. Experimental evaluation confirms that adaptation from the high-noise expert yields superior action-conditioned dynamics modeling compared to the low-noise counterpart.

\begin{figure}
    \centering
    \includegraphics[width=\textwidth]{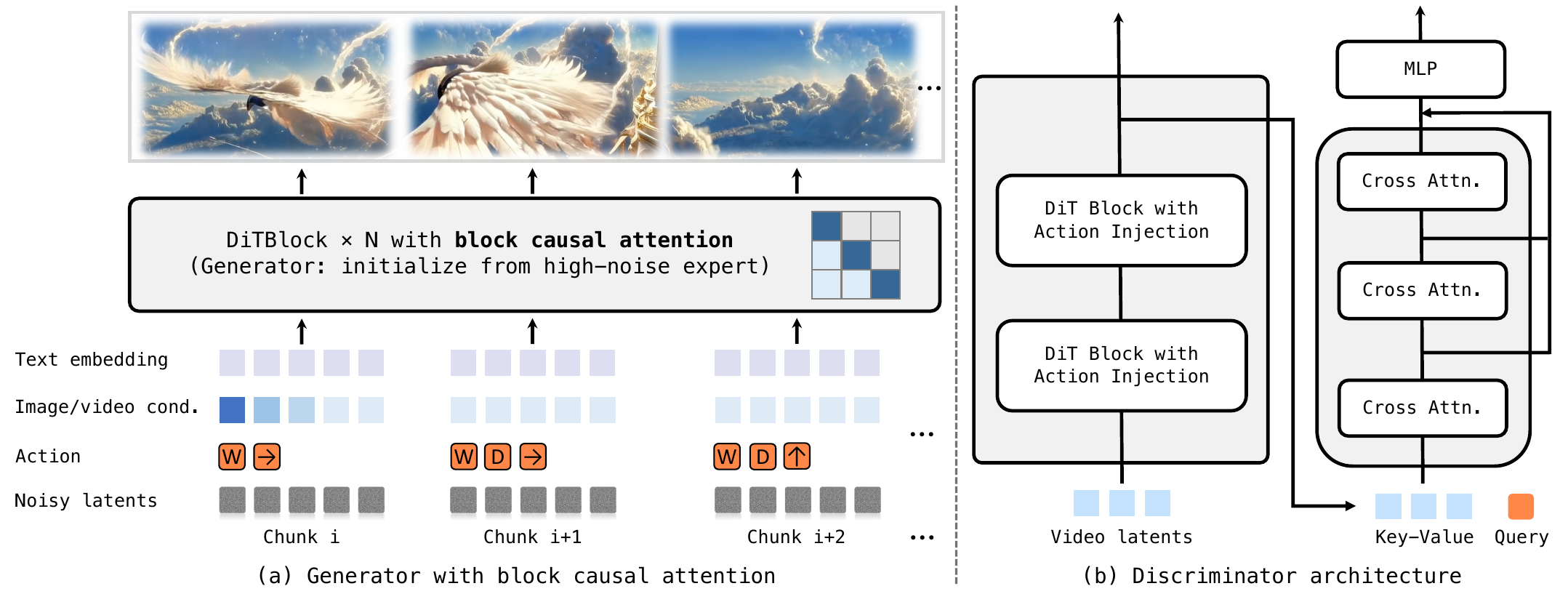}
    \caption{\textbf{(a) Causal generator adaptation.} To enable autoregressive streaming generation, we adapt the high-noise expert using \textbf{block causal attention}. This mechanism enforces global causality across chunks while maintaining local bidirectional consistency for efficient action-conditioned rollouts. \textbf{(b) Discriminator architecture.} For long-horizon training, we attach a GAN classification head $D(\cdot)$ to the fake score network features. This discriminator uses cross-attention to distinguish real from synthesized sequences to mitigate accumulative drift during distribution matching distillation.}
    \label{fig:post-train}
\end{figure}

\noindent\textbf{Architecture adaptation.} 
We adapt the bidirectional teacher into a causal world model following recent autoregressive video distillation frameworks \cite{chen2025skyreels,teng2025magi}. Specifically, we replace full bidirectional temporal attention with block causal attention, which combines local bidirectional attention with global causality constraints to balance modeling capacity with autoregressive requirements \cite{yin2025slow,huang2025self}. 
Within each temporal chunk, tokens attend bidirectionally to capture short-range temporal dependencies and maintain local consistency across neighboring frames. Across chunks, attention is restricted causally such that tokens in the current chunk can only attend to tokens in the same or preceding chunks, eliminating future frame dependencies. This hybrid attention pattern enables unbounded autoregressive generation while preserving long-range temporal coherence.
During inference, the causal structure facilitates efficient streaming generation through key-value caching. We reuse cached representations from previous chunks and compute attention only for newly generated tokens, substantially reducing computational overhead at each generation step.

\noindent\textbf{Training protocol.} 
During training, we process sequences of $N$ noisy video frames partitioned into $L$ chunks, where each chunk is assigned an independent noise timestep following the diffusion forcing paradigm \cite{chen2024diffusion,yin2025slow}. To optimize training efficiency while preserving model expressiveness, we restrict training to a small set of strategically selected target timesteps $\{t_1, \dots, t_m\}$ that serve as distillation targets in subsequent stages. These timesteps are chosen to span the denoising trajectory while maintaining computational tractability.
Since our initialization uses the high-noise expert, which was exclusively trained on high-noise conditions, we augment the training with clean frame supervision by including timestep 0 sampling \cite{hong2025relic,tang2025hunyuan}. This enables the model to learn clean latent encoding capabilities, effectively bridging the specialization gap between the high-noise and low-noise experts from our teacher model. The training loss is formulated as follows, 
\begin{align}
\Loss=\E_{x^i\in p(x), t\in\{t_1, \dots, t_m\}}\left\|G_\theta(x_{t}^i,t,a) - x_0^i\right\|^2,
\end{align}
where $G_\theta$ is the student network, $p(x)$ denotes the distribution of video data, and $a$ is the action condition.

\subsubsection{Few-Step Distillation with Long-Horizon Training}
\label{subsec:dmd}
While our causal adapted model generates visually plausible video frames following user input actions, significant drift accumulates beyond the training horizon due to a distribution mismatch between training and inference conditions. To address this fundamental challenge, we employ a comprehensive distillation framework that combines self-forcing training with advanced distribution matching techniques.

\noindent\textbf{Self-rollout extended horizon training.} 
Following the self-forcing paradigm \cite{huang2025self,lu2025reward,yang2025longlive,apt2}, we train the student model on its own generated sequences to bridge the train-test gap. During training, the model is conditioned on its previously generated frames stored via an efficient rolling key-value cache, forcing it to develop robust recovery mechanisms from its own generation artifacts and accumulated errors. This approach ensures that the model learns to handle the distribution shift that naturally occurs during autoregressive generation.
To manage the substantial computational overhead of long-horizon rollouts, we employ stochastic gradient truncation strategies. Specifically, we backpropagate gradients only through the most recent $K$ generation steps while maintaining the full context for forward computation, balancing training efficiency with long-term dependency learning.

\noindent\textbf{Distribution matching and adversarial optimization.}
We apply distribution matching distillation (DMD) combined with adversarial optimization \cite{yin2024one,yin2024improved} to improve sample quality and temporal consistency. We use the middle-trained MoE teacher model as our real score function and initialize the fake score model using the same MoE teacher for full-step score matching.
For action-conditioned generation, the gradient with respect to student parameters $\theta$ is:

\begin{align}
\nabla_\theta \mathbb{E}_t[D_{\text{KL}}(p_{\theta,t} \| p_{\text{data},t})] = -\mathbb{E}_{t,\hat{x}_t \sim q_{t|0}(\hat{x}_t|\tilde{x}), \tilde{x} \sim p_\theta(\tilde{x}|a)} \left[ (s_{\text{real}}(\hat{x}_t, t, a) - s_{\text{fake}}(\hat{x}_t, t, a)) \frac{\partial \hat{x}}{\partial \theta} \right],
\end{align}
where $p_{\theta,t}$ is the student distribution at timestep $t$, $p_{\text{data},t}$ is the data distribution at $t$,  $\tilde{x}$ are the clean samples generated by the student, $\hat{x}_t$ are the noisy version obtained via forward diffusion, $a$ is the action condition, $s_{\text{real}}$ and $s_{\text{fake}}$ are the approximated scores using the real and fake score networks, respectively. 
This gradient is equivalent to the following tractable optimization objective:

\begin{align}\label{eq:dmd}
\mathcal{L}_{\text{DMD}}(\theta) = \mathbb{E}_{t,\hat{x}_t,\hat{x},a} \left[ \frac{1}{2} \left\| \hat{x} - \text{sg}[\hat{x} - (\mu_{\rm real}(\hat{x}_t, t, a) - \mu_{\rm fake}^\phi(\hat{x}_t, t, a))] \right\|^2 \right],
\end{align}
where $\mu^\phi_{\rm fake}$ represents the fake score network with parameters $\phi$, and $\text{sg}[\cdot]$ denotes the stop-gradient operator.
During DMD training, the student network is updated using~\cref{eq:dmd}. The fake score network $\mu^\phi_{\rm fake}$ is trained with the standard diffusion loss on student-generated videos, while the real score network $\mu_{\rm real}$ is kept fixed. Following~\cite{yin2024improved}, we adopt a two-time-scale update rule: perform multiple updates of $\mu^\phi_{\rm fake}$ for each student update so that $\mu^\phi_{\rm fake}$ closely tracks the student’s evolving output distribution, improving training stability and distillation quality.

However, a performance gap remains between the distilled generator and the teacher model after DMD training; for example, videos produced by the student often exhibit degraded quality. 
Several factors may contribute to this gap.
First, the student is initialized from the high-noise model, and therefore does not inherit the knowledge learned by the low-noise model (i.e., the component responsible for fine details and high-frequency synthesis).
Second, we replace the attention mask with a causal variant and employ only a few sampling steps at inference time, which further limits generation quality.
More importantly, under DMD training, neither the generator nor the teacher is directly supervised by real data, which can cause the student to inherit the teacher’s biases and errors.
To mitigate these issues, we introduce an additional objective based on adversarial training~\cite{gan}. Specifically, the generator aims to fool a discriminator, while the discriminator learns to distinguish real videos from synthesized ones. By incorporating supervision from real data, the distilled generator is no longer strictly bounded by the teacher’s limitations, which can potentially improve sample realism and perceptual quality.

Concretely, we attach a classification head $D(\cdot)$ to the fake score network in DMD. The architecture of the head follows the design in APT~\cite{apt1}. The adversarial objectives are:
\begin{align} 
   \Loss_G & = \E_{p(\tilde{x})}[f(1 - D(\mu_{\rm fake}(\tilde{x}_t, t, a)))], \label{eq:gan-g} \\
   \Loss_D & =  \E_{p(x)}[f(D(\mu_{\rm fake}(x_t, t, a)))] - \E_{p(\tilde{x})}[f(1 - D(\mu_{\rm fake}(\tilde{x}_t, t, a)))], \label{eq:gan-d}
\end{align}
where $p(x)$ and $p(\tilde{x})$ denote the distributions of real and synthesized videos, respectively. $\mu_{\rm fake}$ is the fake score network, $t$ denotes the current denoising timestep in self-forcing~\cite{huang2025self}, and $f(\cdot)$ is the softplus function. Notably, the adversarial loss is used only to update the discriminator head $D$, while the fake score network $\mu_{\rm fake}$ is updated solely with the DMD loss. In addition, we do not apply regularization terms such as R1 or R2~\cite{which2018training}, as the DMD objective is sufficiently stable in our setting.
With this augmented adversarial objective, we substantially improve visual quality while preserving action-following ability and maintaining temporal consistency over long horizons.
\section{Evaluation}\label{sec:eval}

\begin{figure*}[!p]
\centering
\includegraphics[width=\linewidth]{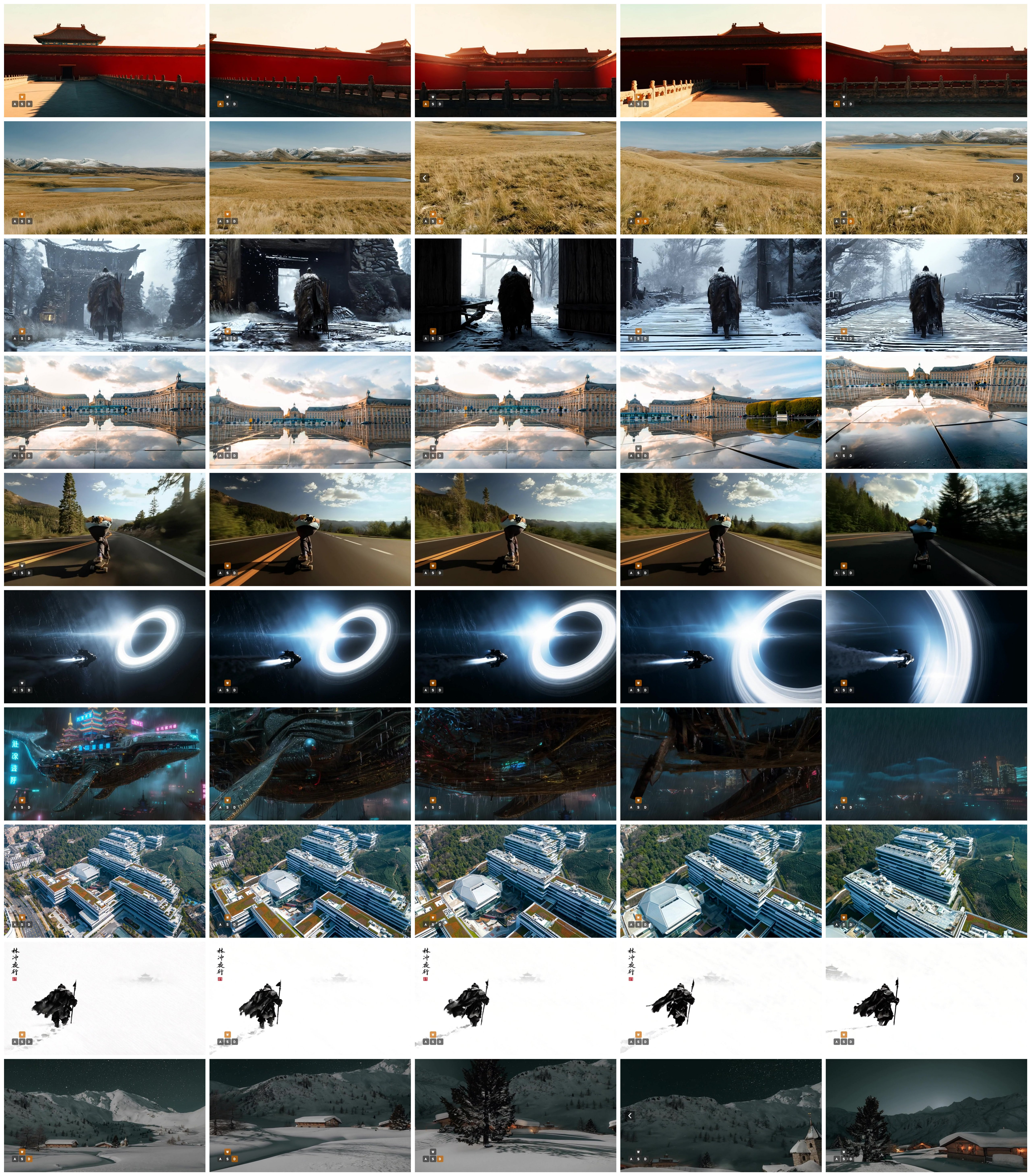}
\caption{\textbf{Qualitative results of} \methodbase\textbf{.}}
\label{fig:qualitative_results_1}
\end{figure*}

\begin{figure*}[!p]
\centering
\includegraphics[width=\linewidth]{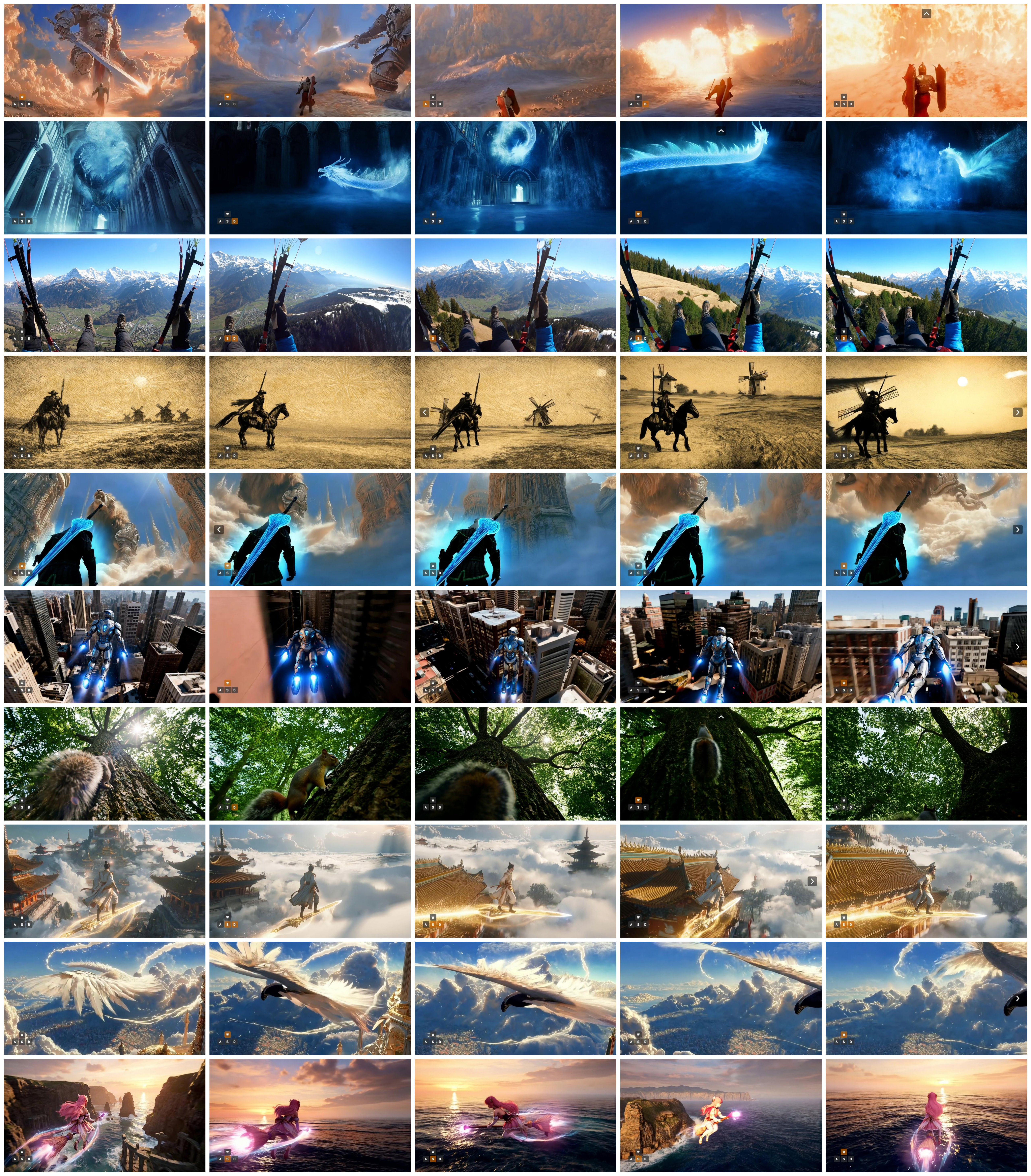}
\caption{\textbf{Qualitative results of} \methodbase\textbf{.}}
\label{fig:qualitative_results_2}
\end{figure*}

\begin{figure*}[!p]
\centering
\includegraphics[width=\linewidth]{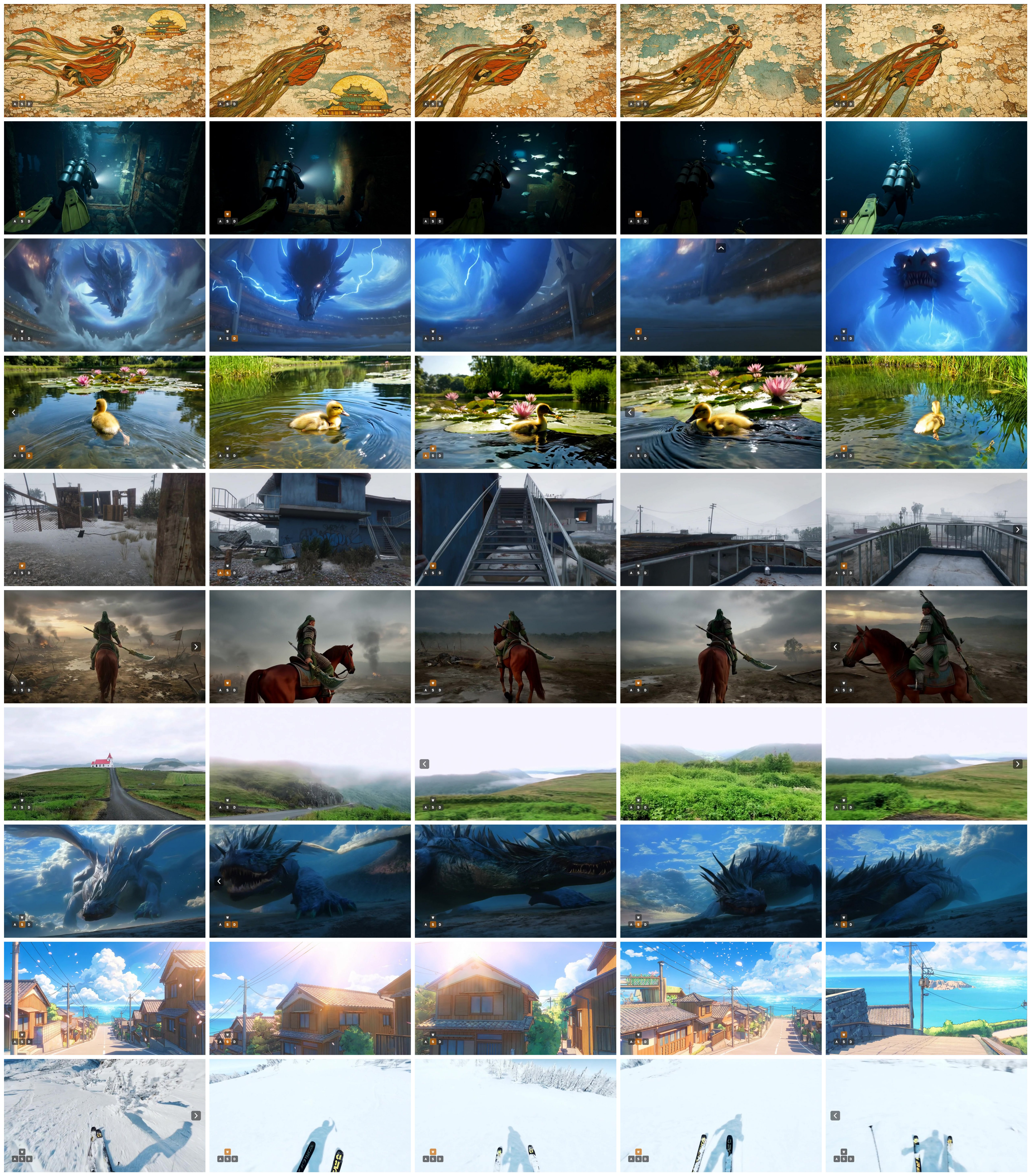}
\caption{\textbf{Qualitative results of} \methodbase\textbf{.}}
\label{fig:qualitative_results_3}
\end{figure*}

\begin{figure*}[!p]
\centering
\includegraphics[width=\linewidth]{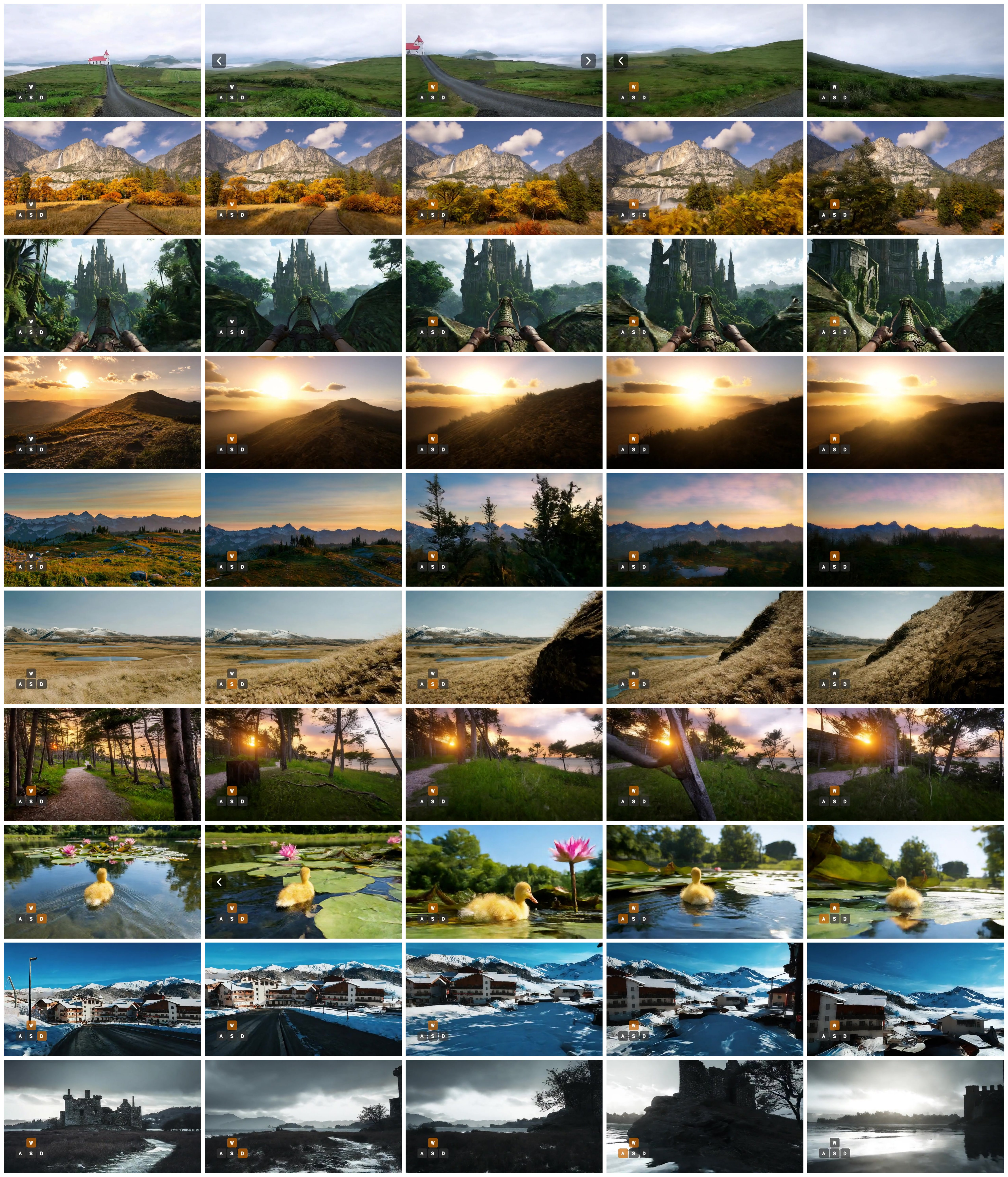}
\caption{\textbf{Qualitative results of} \methodfast\textbf{.}}
\label{fig:qualitative_results_4}
\end{figure*}

\begin{figure*}[!p]
\centering
\includegraphics[width=\linewidth]{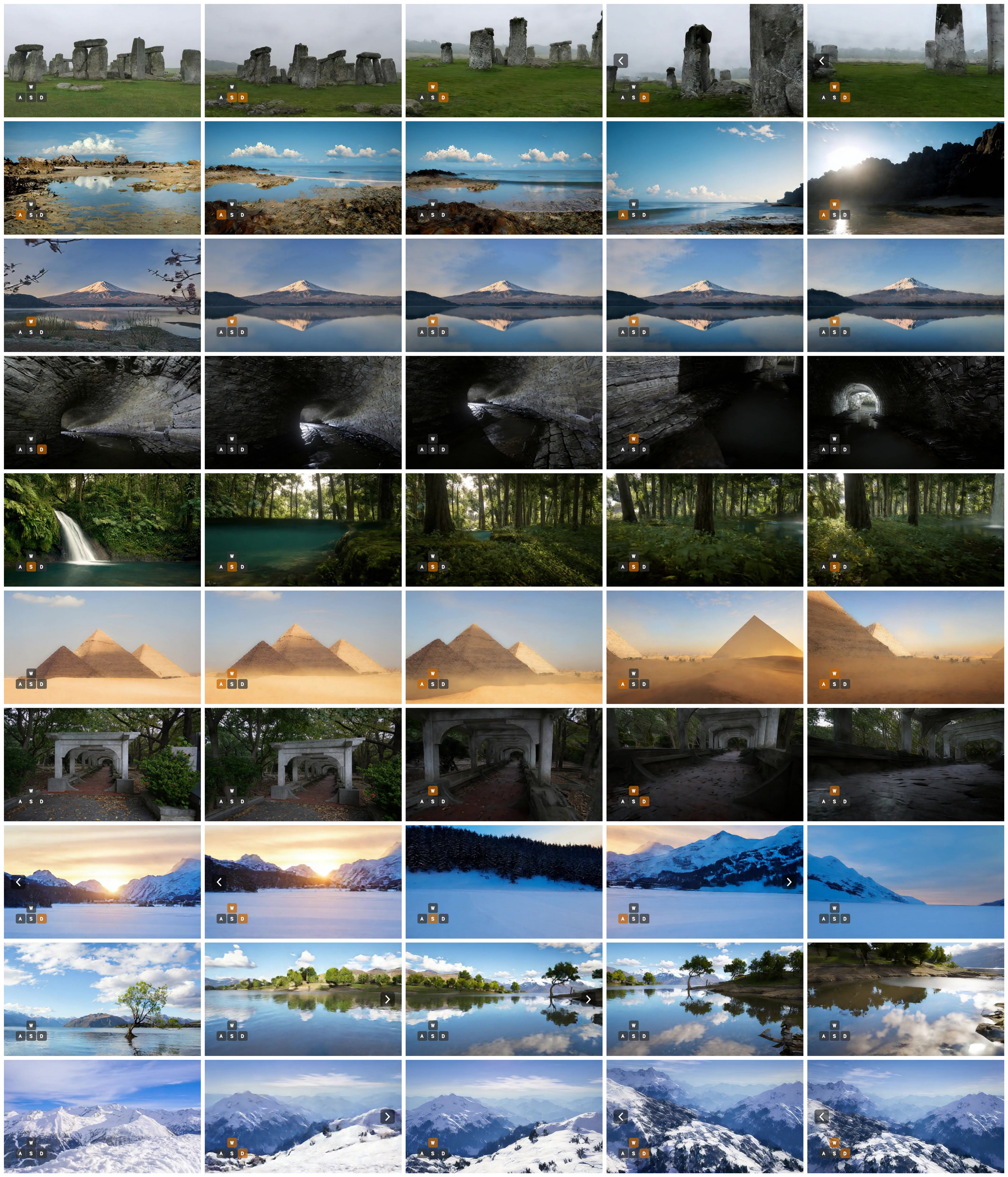}
\caption{\textbf{Qualitative results of} \methodfast\textbf{.}}
\label{fig:qualitative_results_5}
\end{figure*}

\subsection{Qualitative Analysis}
\subsubsection{Diverse Results}
We evaluate the generalization capability of our framework by analyzing the qualitative outcomes of both the middle-trained model \methodbase and the post-trained model \methodfast across a diverse set of scenarios.
\cref{fig:qualitative_results_1,fig:qualitative_results_2,fig:qualitative_results_3} visualize the results from \methodbase, where each row displays keyframes sampled over time.
First, regarding the high-fidelity \methodbase, the sequences demonstrate that it effectively handles varying object properties and complex spatial configurations. The transition between frames remains smooth and logically consistent, highlighting the model's ability to capture fine-grained environmental dynamics.

Building upon this, we further analyze \methodfast, our real-time variant, which achieves 16 fps throughput when processing 480p videos on a system with one GPU node. Although the acceleration process introduces a necessary trade-off in theoretical upper-bound quality, the visual degradation is perceptually marginal. As shown in~\cref{fig:qualitative_results_4,fig:qualitative_results_5}, \methodfast successfully preserves the structural integrity and physical logic of the teacher model. It adapts to dynamic interactions without exhibiting significant visual artifacts or mode collapse, demonstrating that it achieves an optimal balance between inference speed and generation quality.

\begin{figure*}[!t]
\centering
\includegraphics[width=\linewidth]{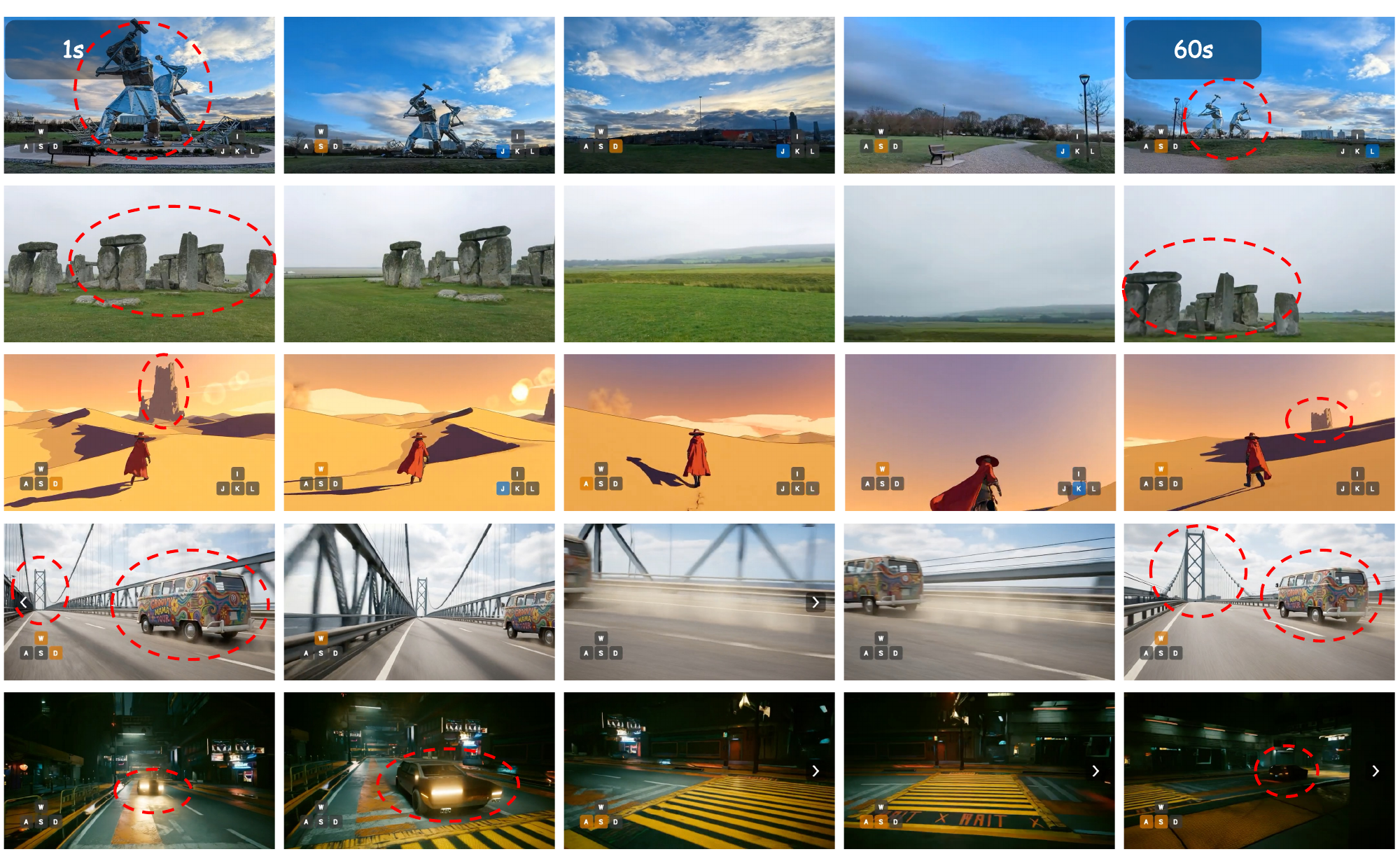}
\caption{\textbf{Emergent memory capability.}
Our model exhibits the emergent ability to maintain long-term consistency and reason about unobserved states. Row 1-3: Static landmarks, such as Stonehenge, preserve their structural integrity even after being out of view for 60 seconds. Row 4-5: The model simulates coherent world dynamics even for unobserved regions: the distant bridge appears closer when the camera returns to the frontal view after moving forward (row 4), and the car continues traveling down the road while out of view (row 5).}
\label{fig:memory}
\end{figure*}

\begin{figure*}[!p]
\centering
\includegraphics[width=\linewidth]{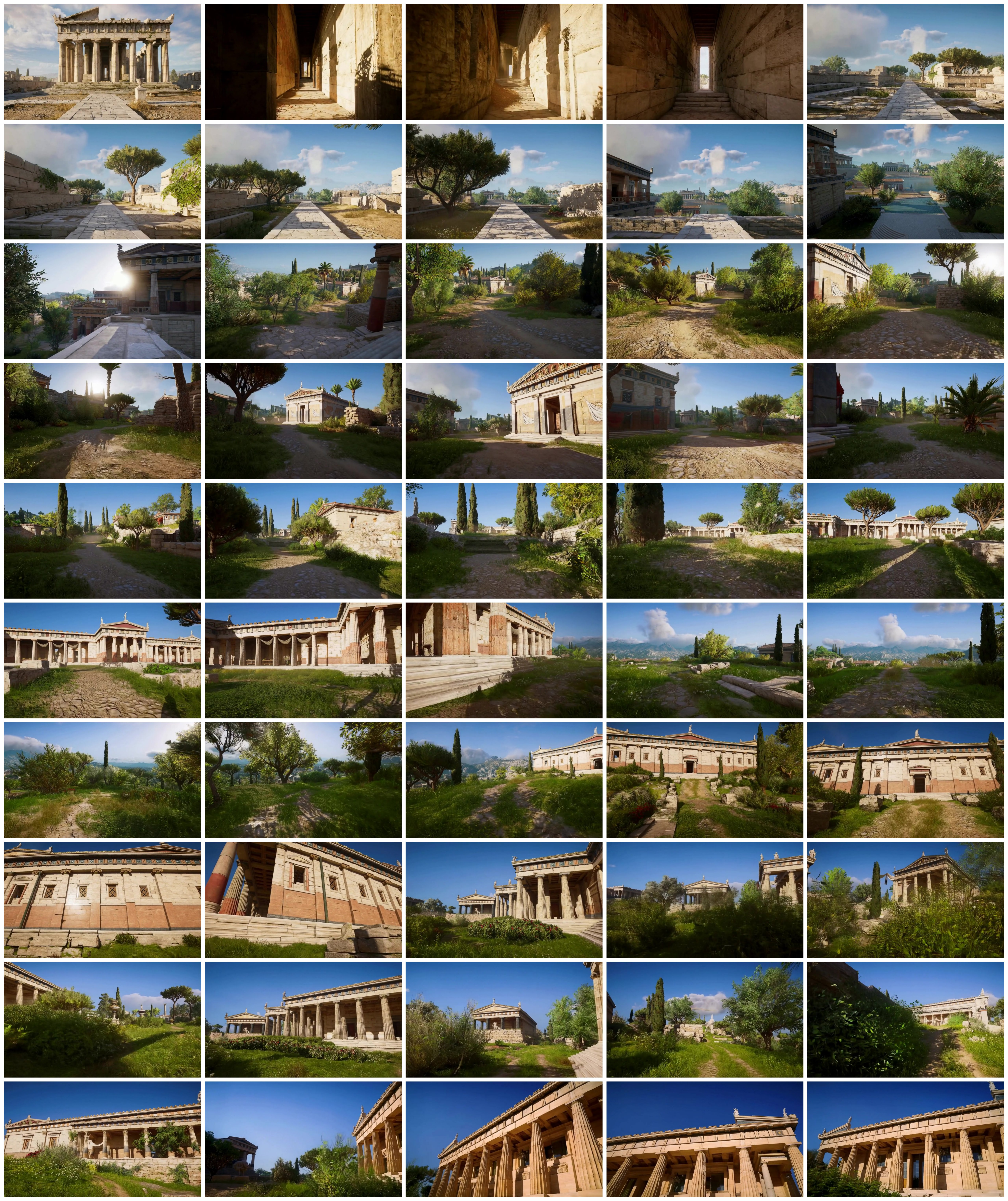}
\caption{\textbf{Ultra-long video generation.} We demonstrate the capability of our model to generate coherent video sequences extending up to 10 minutes in duration.}
\label{fig:qualitative_results_6}
\end{figure*}

\subsubsection{Emergent Memory Capability}

A key property of \method is the emergent ability to maintain global consistency without relying on explicit 3D representations such as Gaussian Splatting~\cite{kerbl20233d}. As shown in the first three rows of~\cref{fig:memory}, the model preserves the structural integrity of landmarks, including statues and Stonehenge, even after they have been out of view for a long duration up to 60 seconds. This aligns with prior observations~\cite{genie3,holocine} that video models possess implicit memory for object reappearance. Crucially, unlike explicit 3D methods that are typically constrained to static scene reconstruction, our video-based approach is far more dynamic. It naturally models complex non-rigid dynamics like flowing water or moving pedestrians that are notoriously difficult for traditional static 3D representations to capture.

Beyond merely rendering visible dynamics, the model even emerges with the capability to reason about the evolution of unobserved states. For instance, as illustrated in row 4 of~\cref{fig:memory}, when the camera returns to a frontal view after moving forward, the distant bridge is rendered significantly closer, accurately reflecting the forward movement over time. Similarly, in row 5, a vehicle leaves the frame, continues its trajectory while unobserved, and reappears at a physically plausible location, rather than vanishing or freezing. These behaviors indicate that the model simulates the underlying spatio-temporal consistency of the real world rather than just memorizing pixels.

\subsubsection{Exploring the Generation Boundary}
As demonstrated in~\cref{fig:qualitative_results_6}, we push the boundaries of temporal coherence in video synthesis. Our model is capable of sustaining stable, high-fidelity environments for ultra-long durations (up to ten minutes) without significant degradation in visual quality or narrative consistency. This result highlights the robustness of our approach in handling long-term temporal dependencies.

\subsection{Quantitative Analysis}

\begin{table}[t]
    \footnotesize
    \centering
    \caption{
        \textbf{Quantitative comparisons.} We compare our model against recent state-of-the-art approaches on VBench~\cite{huang2024vbench}. Our model excels in superior motion dynamics, while maintaining visual quality and temporal smoothness on par with leading competitors.
    }
    \label{tab:quantitative}
    \SetTblrInner{rowsep=1.2pt}      
    \SetTblrInner{colsep=3.2pt}
    \definecolor{bg_purple}{HTML}{A5A3FA}
    \begin{tblr}{
        cells={halign=c,valign=m},
        column{1}={halign=l},
        row{4}={bg=bg_purple,fg=white},
        hline{2}={1-7}{},
        hline{1,5}={1.0pt},
    }
    \textbf{Model} & \textbf{Imaging Quality} & \textbf{Aesthetic Quality} & \textbf{Dynamic Degree} & \textbf{Motion Smooth} & \textbf{Temporal Flickering} & \textbf{Overall Consistency} \\
    Yume-1.5~\cite{mao2025yume15} & 0.5838 & 0.5185 & 0.7612 & 0.9709 & 0.9545 & 0.1994 \\
    HY-World 1.5~\cite{sun2025worldplay} & 0.6512 & 0.5487 & 0.7217 & 0.9897 & 0.9773 & 0.2016 \\
    Ours & 0.6683 & 0.5660 & 0.8857 & 0.9895 & 0.9648 & 0.2178 \\
    \end{tblr}
\end{table}

For quantitative evaluation, considering that evaluation protocols for world models are still in a nascent stage and the proposed method is based on video generative models, we conduct a comprehensive analysis using VBench~\cite{huang2024vbench} on a curated test set comprising 100 generated videos, each exceeding 30 seconds in duration. We compare our \method against two state-of-the-art video world models: Yume-1.5~\cite{mao2025yume15} and HY-World 1.5~\cite{sun2025worldplay}. As shown in~\cref{tab:quantitative}, our method demonstrates superior performance across the majority of evaluated metrics. Specifically, in terms of visual fidelity, our model achieves the highest scores in both \textit{imaging quality} and \textit{aesthetic quality}, surpassing both baselines. This indicates that our model generates scenes with higher realism and better visual appeal, which is essential for an immersive user experience during interactive world roaming.

Crucially for an interactive world model, our model exhibits a significant advantage in \textit{dynamic degree}, achieving a score of 0.8857 compared to 0.7612 for Yume-1.5 and 0.7217 for HY-World 1.5. This substantial margin suggests that our model is capable of generating richer scene transitions and more complex motion in response to user controls, avoiding the static patterns often observed in video generation. Furthermore, despite this high degree of dynamism, our method maintains the best \textit{overall consistency}, confirming that our model maintains strong semantic fidelity to the input prompts throughout long-term generation.

For temporal characteristics, our model achieves competitive results in \textit{motion smoothness} and \textit{temporal flickering}, comparable to the leading baseline HY-World 1.5. This ensures that the generated video streams remain fluid and free from jarring artifacts. In summary, the quantitative results validate that our model not only provides a more dynamic and interactive environment but also maintains superior visual quality and consistency compared to existing approaches.
\section{Applications}\label{sec:application}
\begin{figure*}[!p]
\centering
\includegraphics[width=\linewidth]{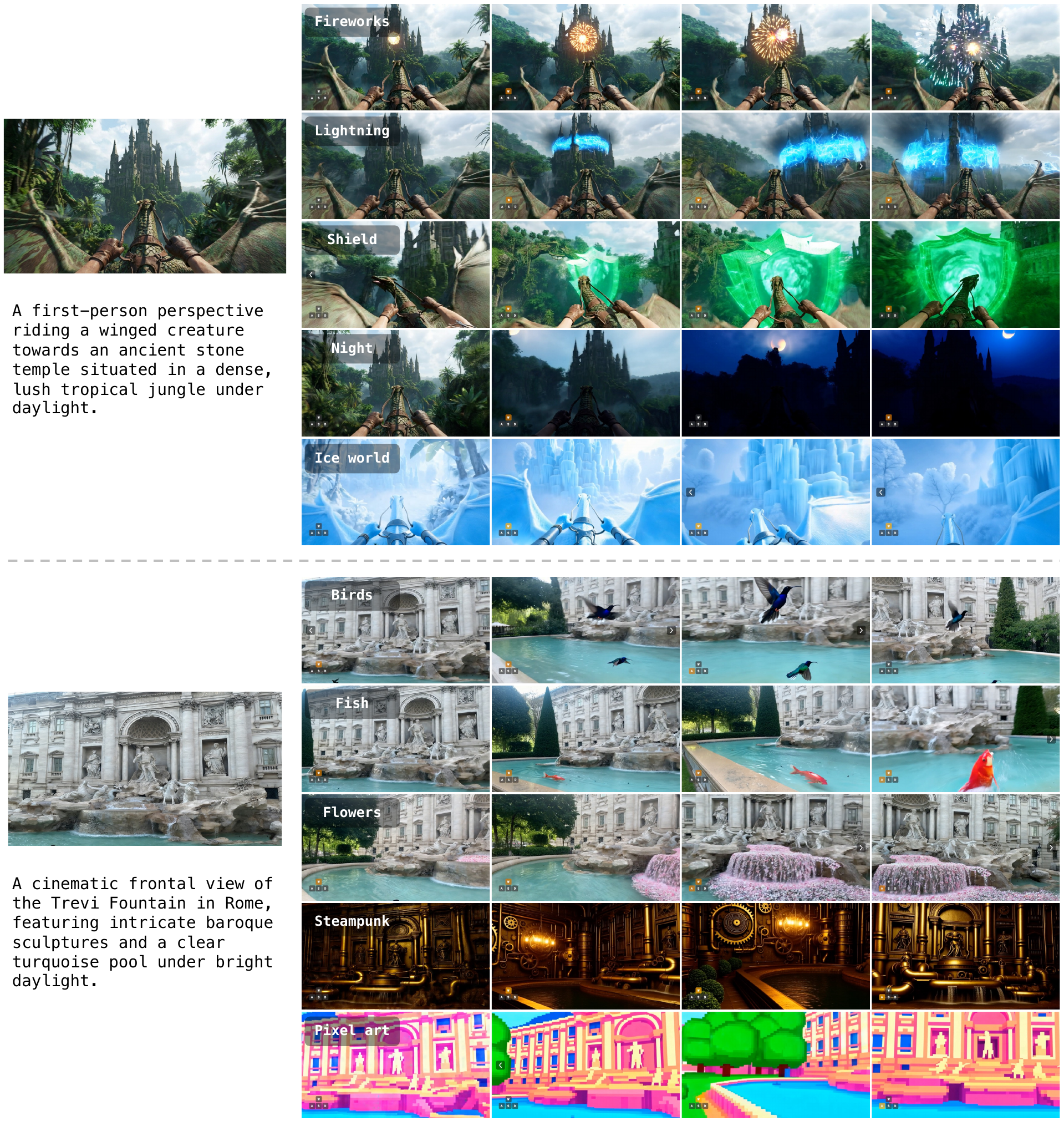}
\caption{\textbf{Promptable world event}. Given a single initial context (left), our model generates diverse future trajectories steered by text prompts. We demonstrate this capability across distinct domains: a fantasy scenario (top) and a realistic scene (bottom). The results highlight our model's ability to handle both \textit{global} environmental shifts (e.g., ``winter'', ``pixel art'') and precise \textit{local} interventions (e.g., ``fireworks'', ``fish''), all while maintaining physical and temporal coherence.}
\label{fig:promptable_world_event}
\end{figure*}

Our autoregressive framework transforms video generation into an interactive simulation by conditioning synthesis on both natural language prompts and discrete actions. This multimodal steerability enables the model to serve as a versatile platform for downstream tasks. In this section, we demonstrate three key applications enabled by our design: (1) \textbf{promptable world events}, where users semantically control global and local dynamics via text; (2) \textbf{action agent}, which leverages the simulator to learn autonomous exploration policies; and (3) \textbf{3D reconstruction}, which validates the emergent geometric consistency and long-term spatial memory of our generated environments.

\begin{figure*}[t]
\centering
\includegraphics[width=\linewidth]{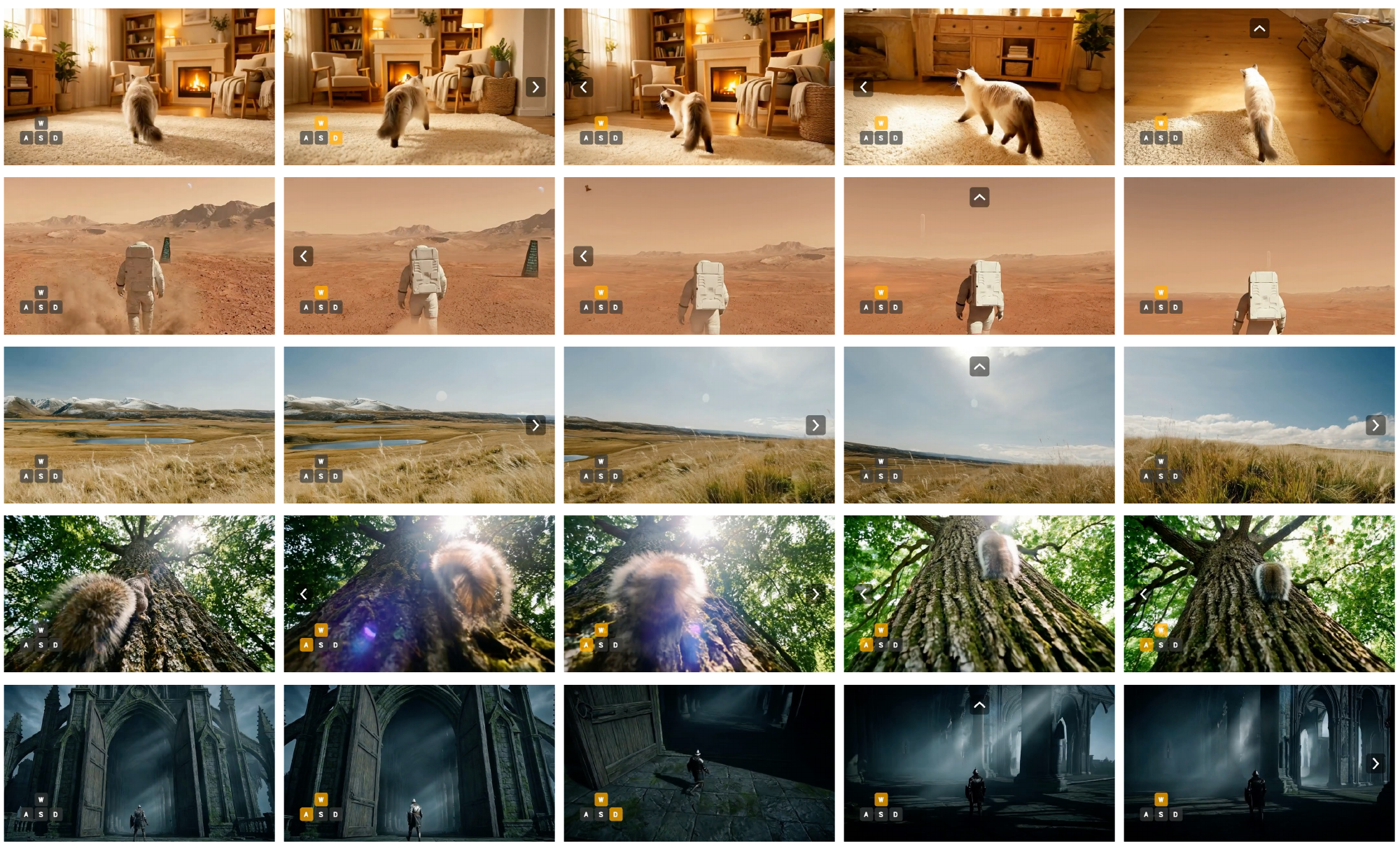}
\caption{\textbf{Application of action agent}.
Given an initial image, the action agent predicts a sequence of actions that simulate exploration in the environment. The predicted actions are converted into camera trajectories, which drive the subsequent world generation.}
\label{fig:agentic-action}
\end{figure*}

\begin{figure}[!p]
\centering
\includegraphics[width=\linewidth]{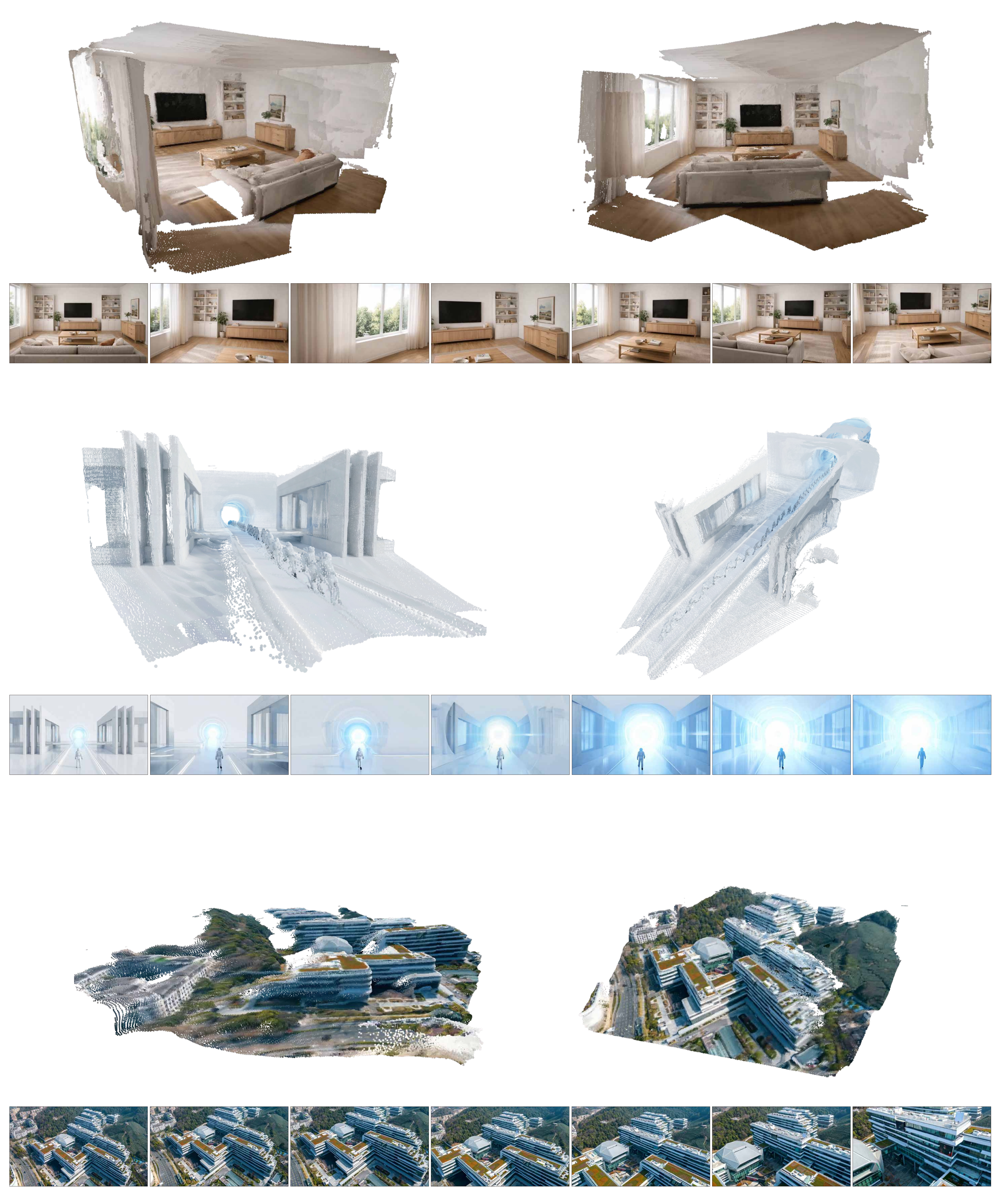}
\caption{\textbf{3D reconstruction results from LingBot-World generated videos.}
Reconstructed point clouds from indoor, sci-fi, and outdoor scenarios demonstrate high spatial consistency and geometric fidelity across diverse environments.
}
\label{fig:app_3drecon}
\end{figure}

\subsection{Promptable World Events}
Instead of restricting users to passive navigation within a static environment, we advocate for a reactive world model where the simulation unfolds differently based on interaction. To this end, we demonstrate poromptable world events~\cite{genie3, worldcanvas}, a mechanism that allows users to actively steer the future trajectory via natural language. As illustrated in~\cref{fig:promptable_world_event}, this capability transforms the generation process from a single deterministic path into a tree of diverse possibilities. Given a single initial context, our model can branch into distinctly different futures based on semantic prompts. This steerability opens up two critical capabilities.

\subsubsection{Global Events}
Global events refer to holistic modifications of the simulation environment, including weather conditions, lighting, and stylistic rendering. Leveraging the text-conditional nature of our base model and the variant of Ditto~\cite{bai2025ditto}, we can manipulate the global state by adjusting the prompts during inference. As illustrated in~\cref{fig:promptable_world_event}, incorporating environment descriptors (e.g., ``winter'' or ``night'') seamlessly transitions the scene into the target domain. The model consistently renders coherent physical effects, such as the freezing of the castle or lighting changes at night, while maintaining temporal consistency with the previous history. Furthermore, the model supports stylistic domain shifts. By prompting for artistic styles (e.g., ``pixel art'' or ``steampunk''), we can transform the visual rendering while preserving the underlying geometry and motion dynamics.

\subsubsection{Local Events}
Local events involve the precise injection of specific objects or dynamic agents into the scene. As shown in~\cref{fig:promptable_world_event}, users can introduce targeted elements, such as triggering ``fireworks'' above a castle or spawning ``birds'' and ``fish'' at a fountain. Our model seamlessly incorporates injected elements into the evolving scene, ensuring physically consistent behavior and temporally stable integration.
This granular control is crucial for embodied AI and autonomous driving. It enables the construction of diverse, interactive training environments where agents must reason about causal relationships and dynamic changes. By defining specific events, we can rigorously evaluate an agent's ability to perceive, predict, and react to fine-grained physical interactions, bridging the gap between static dataset learning and real-world adaptability.

\subsection{Action Agent}

Besides learning an action-conditioned world model, we additionally leverage the same data to train an action agent that infers motion dynamics from single visual observations and incentivizes environment exploration, enabling more effective use of the dataset.

Formally, we fine-tune the Qwen3-VL-2B \cite{bai2025qwen3} backbone on image-action pairs. Each training example consists of a visual observation followed by a sequence of action chunks $(a_0, a_1, \ldots)$, where each $a_i$ specifies the subsequent action that drives the agent to explore the environment. Given the visual observations, the model is trained to predict future actions.

In our setup, the agent outputs the actions for the next 10 seconds, including discrete keyboard controls (\texttt{W}, \texttt{A}, \texttt{S}, \texttt{D}) for locomotion and discretized mouse directions (\texttt{I}, \texttt{J}, \texttt{K}, \texttt{L}) for camera rotations. The predicted actions are then converted into motion trajectories and passed to the world model to generate the corresponding video rollout. Visualizations of the generated result are shown in~\cref{fig:agentic-action}.

\subsection{3D Reconstruction}
Benefiting from high-quality large-scale long-horizon training, \method exhibits an emergent capability of 3D spatial consistency and long-term spatial memory. 
As shown in~\cref{fig:app_3drecon}, by leveraging large-scale 3D reconstruction foundation models~\cite{lin2025depth,wang2025vggt}, we can further convert the generated video sequences into high-quality scene point clouds. 
These point clouds demonstrate strong spatial coherence across frames, serving as a promising source of diverse data for downstream embodied intelligence training. 
Such emergent 3D consistency effectively alleviates the cross-view inconsistency commonly observed in conventional video generation models, thereby enabling superior scene fidelity and geometric accuracy.
\section{Conclusion and Discussion}

\subsection{Summary: A New Open-Source Frontier}
In this report, we present a comprehensive framework that establishes a new open-source frontier for world models, effectively bridging the gap between video generation and actionable simulation. Our contributions cover the entire pipeline, starting with a robust \textbf{data engine} featuring a scalable, automated collection system that ensures high-quality and diverse training data. On the \textbf{modeling} front, we develop a causal transformer architecture optimized for accurate action-control and employ real-time distillation to enable efficient inference. These advancements culminate in diverse \textbf{applications}, demonstrating the model's capability in executing agentic actions, performing consistent world editing, and supporting 3D environment reconstruction.

\subsection{Limitations}
Despite these advancements, several challenges remain in achieving a fully immersive and persistent virtual world. 
\begin{itemize}
    \item \textbf{Memory stability:} Currently, the model's memory is an \textit{emergent ability} derived from the context window rather than an explicit storage module. Consequently, it lacks stability, leading to inconsistencies during long-term simulation.
    \item \textbf{Computational cost:} The inference cost remains high. Running the model requires enterprise-grade GPUs, making it inaccessible to consumer-level hardware.
    \item \textbf{Limited action space:} The range of controllable actions is currently restricted. The model primarily handles navigation and basic movements, lacking a diverse repertoire of complex interactions.
    \item \textbf{Interaction precision:} Fine-grained control remains difficult. Specifically, interacting with a specific target object (e.g., picking up a specific cup on a cluttered table) is challenging due to the lack of precise object-level grounding.
    \item \textbf{Generation length \& drifting:} The coherent generation length is insufficient for extended gameplay. As the video length increases, the scene suffers from ``drifting'' issues, where the environment gradually loses its original structure.
    \item \textbf{Single-agent simulation:} The current framework supports only single-agent perspectives and does not yet account for multi-agent interactions.
\end{itemize}

\subsection{Next Steps}
Looking ahead, we aim to address these limitations through a targeted roadmap. Our primary goal is to significantly expand the action space and enhance the physics engine, allowing for more diverse and realistic interactions with the environment. To solve the stability issue inherent in long-term simulations, we plan to design a better, explicit memory module rather than relying solely on emergent capabilities. Furthermore, we will focus on solving the drifting issue to enable longer video generation, paving the way for infinite-time gameplay and more robust simulations.
\section{Contributors}\label{sec:contributors}

\noindent \textbf{Base Model:} Zelin Gao$^*$, Qiuyu Wang$^*$, Yinghao Xu, Shuailei Ma

\noindent \textbf{Post Training:} Yanhong Zeng$^*$, Jiapeng Zhu$^*$

\noindent \textbf{Games Data:} Ka Leong Cheng$^*$, Yihang Chen, Jie Liu, Yansong Cheng, Yao Yao

\noindent \textbf{Rendering Data:} Yixuan Li$^*$, Jiayi Zhu

\noindent \textbf{Data Pipeline:} Hanlin Wang$^*$, Yihao Meng, Kecheng Zheng

\noindent \textbf{Applications:} Qingyan Bai, Jingye Chen, Zehong Shen, Yue Yu

\noindent \textbf{Project Sponsor:} Xing Zhu, Yujun Shen

\noindent \textbf{Project Lead:} Hao Ouyang

\vspace{5pt}
\noindent \textit{$*$ denotes the leaders of each sub-module.}

\section*{Acknowledgments}
We thank Yu Chen, Zikun Dai, Xiaoyue Duan, Biao Gong, Zhengyu He, Liangxiao Hu, Ting Huang, Bo Jiang, Tao Jiang, Haobo Li, Yangyan Li, Yantao Lin, Fei Lu, Tingzhan Lu, Yunhong Lu, Jianxue Qian, Yipengjing Sun, Jingyun Tian, Yanmeng Wang, Yuanyuan Wang, Yunnan Wang, Leyi Xu, Min Yao, Yufeng Yuan, Han Zhang, Qihang Zhang, Shangzhan Zhang, Shuai Zhou, and Tianxiang Zhou (\textbf{\textit{listed alphabetically by last name}}) for their valuable discussions and assistance.
{
\small
\bibliographystyle{plain}
\bibliography{ref.bib}

\begin{thebibliography}{10}

\bibitem{alonso2024diffusion}
Eloi Alonso, Adam Jelley, Vincent Micheli, Anssi Kanervisto, Amos~J Storkey, Tim Pearce, and Fran{\c{c}}ois Fleuret.
\newblock Diffusion for world modeling: Visual details matter in atari.
\newblock In {\em Adv. Neural Inform. Process. Syst.}, 2024.

\bibitem{assran2025v}
Mido Assran, Adrien Bardes, David Fan, Quentin Garrido, Russell Howes, Mojtaba, Komeili, Matthew Muckley, Ammar Rizvi, Claire Roberts, Koustuv Sinha, Artem Zholus, Sergio Arnaud, Abha Gejji, Ada Martin, Francois~Robert Hogan, Daniel Dugas, Piotr Bojanowski, Vasil Khalidov, Patrick Labatut, Francisco Massa, Marc Szafraniec, Kapil Krishnakumar, Yong Li, Xiaodong Ma, Sarath Chandar, Franziska Meier, Yann LeCun, Michael Rabbat, and Nicolas Ballas.
\newblock V-jepa 2: Self-supervised video models enable understanding, prediction and planning.
\newblock {\em arXiv preprint arXiv:2506.09985}, 2025.

\bibitem{bai2025ditto}
Qingyan Bai, Qiuyu Wang, Hao Ouyang, Yue Yu, Hanlin Wang, Wen Wang, Ka~Leong Cheng, Shuailei Ma, Yanhong Zeng, Zichen Liu, Yinghao Xu, Yujun Shen, and Qifeng Chen.
\newblock Scaling instruction-based video editing with a high-quality synthetic dataset.
\newblock {\em arXiv preprint arXiv:2510.15742}, 2025.

\bibitem{bain2021frozen}
Max Bain, Arsha Nagrani, G{\"u}l Varol, and Andrew Zisserman.
\newblock Frozen in time: A joint video and image encoder for end-to-end retrieval.
\newblock In {\em Int. Conf. Comput. Vis.}, 2021.

\bibitem{genie3}
Philip~J. Ball, Jakob Bauer, Frank Belletti, Bethanie Brownfield, Ariel Ephrat, Shlomi Fruchter, Agrim Gupta, Kristian Holsheimer, Aleksander Holynski, Jiri Hron, Christos Kaplanis, Marjorie Limont, Matt McGill, Yanko Oliveira, Jack Parker-Holder, Frank Perbet, Guy Scully, Jeremy Shar, Stephen Spencer, Omer Tov, Ruben Villegas, Emma Wang, Jessica Yung, Cip Baetu, Jordi Berbel, David Bridson, Jake Bruce, Gavin Buttimore, Sarah Chakera, Bilva Chandra, Paul Collins, Alex Cullum, Bogdan Damoc, Vibha Dasagi, Maxime Gazeau, Charles Gbadamosi, Woohyun Han, Ed~Hirst, Ashyana Kachra, Lucie Kerley, Kristian Kjems, Eva Knoepfel, Vika Koriakin, Jessica Lo, Cong Lu, Zeb Mehring, Alex Moufarek, Henna Nandwani, Valeria Oliveira, Fabio Pardo, Jane Park, Andrew Pierson, Ben Poole, Helen Ran, Tim Salimans, Manuel Sanchez, Igor Saprykin, Amy Shen, Sailesh Sidhwani, Duncan Smith, Joe Stanton, Hamish Tomlinson, Dimple Vijaykumar, Luyu Wang, Piers Wingfield, Nat Wong, Keyang Xu, Christopher Yew, Nick Young, Vadim Zubov, Douglas
  Eck, Dumitru Erhan, Koray Kavukcuoglu, Demis Hassabis, Zoubin Gharamani, Raia Hadsell, A{\"a}ron van~den Oord, Inbar Mosseri, Adrian Bolton, Satinder Singh, and Tim Rockt{\"a}schel.
\newblock Genie 3: A new frontier for world models, 2025.

\bibitem{bar2025navigation}
Amir Bar, Gaoyue Zhou, Danny Tran, Trevor Darrell, and Yann LeCun.
\newblock Navigation world models.
\newblock In {\em IEEE Conf. Comput. Vis. Pattern Recog.}, 2025.

\bibitem{bar2024lumiere}
Omer Bar-Tal, Hila Chefer, Omer Tov, Charles Herrmann, Roni Paiss, Shiran Zada, Ariel Ephrat, Junhwa Hur, Guanghui Liu, Amit Raj, Yuanzhen Li, Michael Rubinstein, Tomer Michaeli, Oliver Wang, Deqing Sun, Tali Dekel, and Inbar Mosseri.
\newblock Lumiere: A space-time diffusion model for video generation.
\newblock In {\em SIGGRAPH Asia}, 2024.

\bibitem{blattmann2023stable}
Andreas Blattmann, Tim Dockhorn, Sumith Kulal, Daniel Mendelevitch, Maciej Kilian, Dominik Lorenz, Yam Levi, Zion English, Vikram Voleti, Adam Letts, Varun Jampani, and Robin Rombach.
\newblock Stable video diffusion: Scaling latent video diffusion models to large datasets.
\newblock {\em arXiv preprint arXiv:2311.15127}, 2023.

\bibitem{brooks2024video}
Tim Brooks, Bill Peebles, Connor Holmes, Will DePue, Yufei Guo, Li~Jing, David Schnurr, Joe Taylor, Troy Luhman, Eric Luhman, Clarence Ng, Ricky Wang, and Aditya Ramesh.
\newblock Video generation models as world simulators.
\newblock {\em OpenAI Blog}, 2024.

\bibitem{bruce2024genie}
Jake Bruce, Michael~D Dennis, Ashley Edwards, Jack Parker-Holder, Yuge Shi, Edward Hughes, Matthew Lai, Aditi Mavalankar, Richie Steigerwald, Chris Apps, et~al.
\newblock Genie: Generative interactive environments.
\newblock In {\em Int. Conf. Mach. Learn.}, 2024.

\bibitem{pyscenedetect}
Brandon Castellano.
\newblock Pyscenedetect: An open-source video scene detection program and python library.
\newblock \url{https://github.com/Breakthrough/PySceneDetect}, 2018.

\bibitem{chen2024diffusion}
Boyuan Chen, Diego Mart{\'\i}~Mons{\'o}, Yilun Du, Max Simchowitz, Russ Tedrake, and Vincent Sitzmann.
\newblock Diffusion forcing: Next-token prediction meets full-sequence diffusion.
\newblock In {\em Adv. Neural Inform. Process. Syst.}, 2024.

\bibitem{chen2025vl}
Delong Chen, Mustafa Shukor, Theo Moutakanni, Willy Chung, Jade Yu, Tejaswi Kasarla, Allen Bolourchi, Yann LeCun, and Pascale Fung.
\newblock Vl-jepa: Joint embedding predictive architecture for vision-language.
\newblock {\em arXiv preprint arXiv:2512.10942}, 2025.

\bibitem{chen2025skyreels}
Guibin Chen, Dixuan Lin, Jiangping Yang, Chunze Lin, Junchen Zhu, Mingyuan Fan, Hao Zhang, Sheng Chen, Zheng Chen, Chengcheng Ma, Weiming Xiong, Wei Wang, Nuo Pang, Kang Kang, Zhiheng Xu, Yuzhe Jin, Yupeng Liang, Yubing Song, Peng Zhao, Boyuan Xu, Di~Qiu, Debang Li, Zhengcong Fei, Yang Li, and Yahui Zhou.
\newblock Skyreels-v2: Infinite-length film generative model.
\newblock {\em arXiv preprint arXiv:2504.13074}, 2025.

\bibitem{chen2024sharegpt4video}
Lin Chen, Xilin Wei, Jinsong Li, Xiaoyi Dong, Pan Zhang, Yuhang Zang, Zehui Chen, Haodong Duan, Bin Lin, Zhenyu Tang, Li~Yuan, Yu~Qiao, Dahua Lin, Feng Zhao, and Jiaqi Wang.
\newblock Sharegpt4video: Improving video understanding and generation with better captions.
\newblock In {\em Adv. Neural Inform. Process. Syst.}, 2024.

\bibitem{chen2024panda}
Tsai-Shien Chen, Aliaksandr Siarohin, Willi Menapace, Ekaterina Deyneka, Hsiang-wei Chao, Byung~Eun Jeon, Yuwei Fang, Hsin-Ying Lee, Jian Ren, Ming-Hsuan Yang, et~al.
\newblock Panda-70m: Captioning 70m videos with multiple cross-modality teachers.
\newblock In {\em IEEE Conf. Comput. Vis. Pattern Recog.}, 2024.

\bibitem{damen2018scaling}
Dima Damen, Hazel Doughty, Giovanni~Maria Farinella, Sanja Fidler, Antonino Furnari, Evangelos Kazakos, Davide Moltisanti, Jonathan Munro, Toby Perrett, Will Price, and Michael Wray.
\newblock Scaling egocentric vision: The epic-kitchens dataset.
\newblock In {\em Eur. Conf. Comput. Vis.}, 2018.

\bibitem{ue}
{Epic Games}.
\newblock {Unreal Engine}.
\newblock \url{https://www.unrealengine.com/}, 2023.
\newblock Accessed: 2026-01-25.

\bibitem{fedus2022switch}
William Fedus, Barret Zoph, and Noam Shazeer.
\newblock Switch transformers: Scaling to trillion parameter models with simple and efficient sparsity.
\newblock {\em JMLR}, 2022.

\bibitem{feng2025survey}
Tuo Feng, Wenguan Wang, and Yi~Yang.
\newblock A survey of world models for autonomous driving.
\newblock {\em arXiv preprint arXiv:2501.11260}, 2025.

\bibitem{gan}
Ian Goodfellow, Jean Pouget-Abadie, Mehdi Mirza, Bing Xu, David Warde-Farley, Sherjil Ozair, Aaron Courville, and Yoshua Bengio.
\newblock Generative adversarial nets.
\newblock In {\em Adv. Neural Inform. Process. Syst.}, 2014.

\bibitem{grauman2022ego4d}
Kristen Grauman, Andrew Westbury, Eugene Byrne, Zachary Chavis, Antonino Furnari, Rohit Girdhar, Jackson Hamburger, Hao Jiang, Miao Liu, Xingyu Liu, Miguel Martin, Tushar Nagarajan, Ilija Radosavovic, Santhosh~Kumar Ramakrishnan, Fiona Ryan, Jayant Sharma, Michael Wray, Mengmeng Xu, Eric~Zhongcong Xu, Chen Zhao, Siddhant Bansal, Dhruv Batra, Vincent Cartillier, Sean Crane, Tien Do, Morrie Doulaty, Akshay Erapalli, Christoph Feichtenhofer, Adriano Fragomeni, Qichen Fu, Abrham Gebreselasie, Cristina Gonzalez, James Hillis, Xuhua Huang, Yifei Huang, Wenqi Jia, Weslie Khoo, Jachym Kolar, Satwik Kottur, Anurag Kumar, Federico Landini, Chao Li, Yanghao Li, Zhenqiang Li, Karttikeya Mangalam, Raghava Modhugu, Jonathan Munro, Tullie Murrell, Takumi Nishiyasu, Will Price, Paola~Ruiz Puentes, Merey Ramazanova, Leda Sari, Kiran Somasundaram, Audrey Southerland, Yusuke Sugano, Ruijie Tao, Minh Vo, Yuchen Wang, Xindi Wu, Takuma Yagi, Ziwei Zhao, Yunyi Zhu, Pablo Arbelaez, David Crandall, Dima Damen, Giovanni~Maria
  Farinella, Christian Fuegen, Bernard Ghanem, Vamsi~Krishna Ithapu, C.~V. Jawahar, Hanbyul Joo, Kris Kitani, Haizhou Li, Richard Newcombe, Aude Oliva, Hyun~Soo Park, James~M. Rehg, Yoichi Sato, Jianbo Shi, Mike~Zheng Shou, Antonio Torralba, Lorenzo Torresani, Mingfei Yan, and Jitendra Malik.
\newblock Ego4d: Around the world in 3,000 hours of egocentric video.
\newblock In {\em IEEE Conf. Comput. Vis. Pattern Recog.}, 2022.

\bibitem{gupta2024photorealistic}
Agrim Gupta, Lijun Yu, Kihyuk Sohn, Xiuye Gu, Meera Hahn, Fei-Fei Li, Irfan Essa, Lu~Jiang, and Jos{\'e} Lezama.
\newblock Photorealistic video generation with diffusion models.
\newblock In {\em Eur. Conf. Comput. Vis.}, 2024.

\bibitem{ha2018world}
David Ha and J{\"u}rgen Schmidhuber.
\newblock World models.
\newblock {\em arXiv preprint arXiv:1803.10122}, 2018.

\bibitem{hacohen2024ltx}
Yoav HaCohen, Nisan Chiprut, Benny Brazowski, Daniel Shalem, Dudu Moshe, Eitan Richardson, Eran Levin, Guy Shiran, Nir Zabari, Ori Gordon, Poriya Panet, Sapir Weissbuch, Victor Kulikov, Yaki Bitterman, Zeev Melumian, and Ofir Bibi.
\newblock Ltx-video: Realtime video latent diffusion.
\newblock {\em arXiv preprint arXiv:2501.00103}, 2024.

\bibitem{hafner2023mastering}
Danijar Hafner, Jurgis Pasukonis, Jimmy Ba, and Timothy Lillicrap.
\newblock Mastering diverse domains through world models.
\newblock {\em arXiv preprint arXiv:2301.04104}, 2023.

\bibitem{he2025matrix}
Xianglong He, Chunli Peng, Zexiang Liu, Boyang Wang, Yifan Zhang, Qi~Cui, Fei Kang, Biao Jiang, Mengyin An, Yangyang Ren, Baixin Xu, Hao-Xiang Guo, Kaixiong Gong, Size Wu, Wei Li, Xuchen Song, Yang Liu, Yangguang Li, and Yahui Zhou.
\newblock Matrix-game 2.0: An open-source real-time and streaming interactive world model.
\newblock {\em arXiv preprint arXiv:2508.13009}, 2025.

\bibitem{hong2025relic}
Yicong Hong, Yiqun Mei, Chongjian Ge, Yiran Xu, Yang Zhou, Sai Bi, Yannick Hold-Geoffroy, Mike Roberts, Matthew Fisher, Eli Shechtman, Kalyan Sunkavalli, Feng Liu, Zhengqi Li, and Hao Tan.
\newblock Relic: Interactive video world model with long-horizon memory.
\newblock {\em arXiv preprint arXiv:2512.04040}, 2025.

\bibitem{huang2023voxposer}
Wenlong Huang, Chen Wang, Ruohan Zhang, Yunzhu Li, Jiajun Wu, and Li~Fei-Fei.
\newblock Voxposer: Composable 3d value maps for robotic manipulation with language models.
\newblock {\em arXiv preprint arXiv:2307.05973}, 2023.

\bibitem{huang2025self}
Xun Huang, Zhengqi Li, Guande He, Mingyuan Zhou, and Eli Shechtman.
\newblock Self forcing: Bridging the train-test gap in autoregressive video diffusion.
\newblock {\em arXiv preprint arXiv:2506.08009}, 2025.

\bibitem{huang2024vbench}
Ziqi Huang, Yinan He, Jiashuo Yu, Fan Zhang, Chenyang Si, Yuming Jiang, Yuanhan Zhang, Tianxing Wu, Qingyang Jin, Nattapol Chanpaisit, Yaohui Wang, Xinyuan Chen, Limin Wang, Dahua Lin, Yu~Qiao, and Ziwei Liu.
\newblock Vbench: Comprehensive benchmark suite for video generative models.
\newblock In {\em IEEE Conf. Comput. Vis. Pattern Recog.}, 2024.

\bibitem{jacobs2023deepspeed}
Sam~Ade Jacobs, Masahiro Tanaka, Chengming Zhang, Minjia Zhang, Shuaiwen~Leon Song, Samyam Rajbhandari, and Yuxiong He.
\newblock Deepspeed ulysses: System optimizations for enabling training of extreme long sequence transformer models.
\newblock {\em arXiv preprint arXiv:2309.14509}, 2023.

\bibitem{kendall2015posenet}
Alex Kendall, Matthew Grimes, and Roberto Cipolla.
\newblock Posenet: A convolutional network for real-time 6-dof camera relocalization.
\newblock In {\em Int. Conf. Comput. Vis.}, 2015.

\bibitem{kerbl20233d}
Bernhard Kerbl, Georgios Kopanas, Thomas Leimk{\"u}hler, and George Drettakis.
\newblock 3d gaussian splatting for real-time radiance field rendering.
\newblock {\em ACM Trans. Graph.}, 2023.

\bibitem{lecun2022path}
Yann LeCun.
\newblock A path towards autonomous machine intelligence version 0.9. 2, 2022-06-27.
\newblock {\em Open Review}, 2022.

\bibitem{lepikhin2020gshard}
Dmitry Lepikhin, HyoukJoong Lee, Yuanzhong Xu, Dehao Chen, Orhan Firat, Yanping Huang, Maxim Krikun, Noam Shazeer, and Zhifeng Chen.
\newblock Gshard: Scaling giant models with conditional computation and automatic sharding.
\newblock {\em arXiv preprint arXiv:2006.16668}, 2020.

\bibitem{megasam}
Zhengqi Li, Richard Tucker, Forrester Cole, Qianqian Wang, Linyi Jin, Vickie Ye, Angjoo Kanazawa, Aleksander Holynski, and Noah Snavely.
\newblock {MegaSaM: Accurate, Fast and Robust Structure and Motion from Casual Dynamic Videos}.
\newblock In {\em IEEE Conf. Comput. Vis. Pattern Recog.}, 2025.

\bibitem{lin2025depth}
Haotong Lin, Sili Chen, Junhao Liew, Donny~Y Chen, Zhenyu Li, Guang Shi, Jiashi Feng, and Bingyi Kang.
\newblock Depth anything 3: Recovering the visual space from any views.
\newblock {\em arXiv preprint arXiv:2511.10647}, 2025.

\bibitem{apt1}
Shanchuan Lin, Xin Xia, Yuxi Ren, Ceyuan Yang, Xuefeng Xiao, and Lu~Jiang.
\newblock Diffusion adversarial post-training for one-step video generation.
\newblock In {\em Int. Conf. Mach. Learn.}, 2025.

\bibitem{apt2}
Shanchuan Lin, Ceyuan Yang, Hao He, Jianwen Jiang, Yuxi Ren, Xin Xia, Yang Zhao, Xuefeng Xiao, and Lu~Jiang.
\newblock Autoregressive adversarial post-training for real-time interactive video generation.
\newblock {\em arXiv preprint arXiv:2506.09350}, 2025.

\bibitem{long2025survey}
Xiaoxiao Long, Qingrui Zhao, Kaiwen Zhang, Zihao Zhang, Dingrui Wang, Yumeng Liu, Zhengjie Shu, Yi~Lu, Shouzheng Wang, Xinzhe Wei, Wei Li, Wei Yin, Yao Yao, Jia Pan, Qiu Shen, Ruigang Yang, Xun Cao, and Qionghai Dai.
\newblock A survey: Learning embodied intelligence from physical simulators and world models.
\newblock {\em arXiv preprint arXiv:2507.00917}, 2025.

\bibitem{lu2024deepseek}
Haoyu Lu, Wen Liu, Bo~Zhang, Bingxuan Wang, Kai Dong, Bo~Liu, Jingxiang Sun, Tongzheng Ren, Zhuoshu Li, Hao Yang, Yaofeng Sun, Chengqi Deng, Hanwei Xu, Zhenda Xie, and Chong Ruan.
\newblock Deepseek-vl: towards real-world vision-language understanding.
\newblock {\em arXiv preprint arXiv:2403.05525}, 2024.

\bibitem{lu2025reward}
Yunhong Lu, Yanhong Zeng, Haobo Li, Hao Ouyang, Qiuyu Wang, Ka~Leong Cheng, Jiapeng Zhu, Hengyuan Cao, Zhipeng Zhang, Xing Zhu, Yujun Shen, and Min Zhang.
\newblock Reward forcing: Efficient streaming video generation with rewarded distribution matching distillation.
\newblock {\em arXiv preprint arXiv:2512.04678}, 2025.

\bibitem{luo2023latent}
Simian Luo, Yiqin Tan, Longbo Huang, Jian Li, and Hang Zhao.
\newblock Latent consistency models: Synthesizing high-resolution images with few-step inference.
\newblock {\em arXiv preprint arXiv:2310.04378}, 2023.

\bibitem{mao2025yume15}
Xiaofeng Mao, Zhen Li, Chuanhao Li, Xiaojie Xu, Kaining Ying, Tong He, Jiangmiao Pang, Yu~Qiao, and Kaipeng Zhang.
\newblock Yume-1.5: A text-controlled interactive world generation model.
\newblock {\em arXiv preprint arXiv:2512.22096}, 2025.

\bibitem{holocine}
Yihao Meng, Hao Ouyang, Yue Yu, Qiuyu Wang, Wen Wang, Ka~Leong Cheng, Hanlin Wang, Yixuan Li, Cheng Chen, Yanhong Zeng, Yujun Shen, and Huamin Qu.
\newblock Holocine: Holistic generation of cinematic multi-shot long video narratives.
\newblock {\em arXiv preprint arXiv:2510.20822}, 2025.

\bibitem{which2018training}
Lars Mescheder, Andreas Geiger, and Sebastian Nowozin.
\newblock Which training methods for {GAN}s do actually converge?
\newblock In {\em Int. Conf. Mach. Learn.}, 2018.

\bibitem{mialon2023gaia}
Gr{\'e}goire Mialon, Cl{\'e}mentine Fourrier, Thomas Wolf, Yann LeCun, and Thomas Scialom.
\newblock Gaia: a benchmark for general ai assistants.
\newblock In {\em Int. Conf. Learn. Represent.}, 2023.

\bibitem{dxc}
Microsoft.
\newblock Directx shader compiler.
\newblock \url{https://github.com/microsoft/DirectXShaderCompiler}, 2017.
\newblock Accessed: 2026-01-25.

\bibitem{mildenhall2021nerf}
Ben Mildenhall, Pratul~P Srinivasan, Matthew Tancik, Jonathan~T Barron, Ravi Ramamoorthi, and Ren Ng.
\newblock Nerf: Representing scenes as neural radiance fields for view synthesis.
\newblock {\em Communications of the ACM}, 2021.

\bibitem{mu2025comprehensive}
Siyuan Mu and Sen Lin.
\newblock A comprehensive survey of mixture-of-experts: Algorithms, theory, and applications.
\newblock {\em arXiv preprint arXiv:2503.07137}, 2025.

\bibitem{agarwal2025cosmos}
NVIDIA.
\newblock Cosmos world foundation model platform for physical ai.
\newblock {\em arXiv preprint arXiv:2501.03575}, 2025.

\bibitem{ali2025world}
NVIDIA.
\newblock World simulation with video foundation models for physical ai.
\newblock {\em arXiv preprint arXiv:2511.00062}, 2025.

\bibitem{achiam2023gpt}
OpenAI.
\newblock {GPT-4} technical report.
\newblock {\em arXiv preprint arXiv:2303.08774}, 2023.

\bibitem{peebles2023scalable}
William Peebles and Saining Xie.
\newblock Scalable diffusion models with transformers.
\newblock In {\em Int. Conf. Comput. Vis.}, 2023.

\bibitem{radford2021learning}
Alec Radford, Jong~Wook Kim, Chris Hallacy, Aditya Ramesh, Gabriel Goh, Sandhini Agarwal, Girish Sastry, Amanda Askell, Pamela Mishkin, Jack Clark, Gretchen Krueger, and Ilya Sutskever.
\newblock Learning transferable visual models from natural language supervision.
\newblock In {\em Int. Conf. Mach. Learn.}, 2021.

\bibitem{ren2025cosmos}
Xuanchi Ren, Yifan Lu, Tianshi Cao, Ruiyuan Gao, Shengyu Huang, Amirmojtaba Sabour, Tianchang Shen, Tobias Pfaff, Jay~Zhangjie Wu, Runjian Chen, Seung~Wook Kim, Jun Gao, Laura Leal-Taixe, Mike Chen, Sanja Fidler, and Huan Ling.
\newblock Cosmos-drive-dreams: Scalable synthetic driving data generation with world foundation models.
\newblock {\em arXiv preprint arXiv:2506.09042}, 2025.

\bibitem{russell2025gaia}
Lloyd Russell, Anthony Hu, Lorenzo Bertoni, George Fedoseev, Jamie Shotton, Elahe Arani, and Gianluca Corrado.
\newblock Gaia-2: A controllable multi-view generative world model for autonomous driving.
\newblock {\em arXiv preprint arXiv:2503.20523}, 2025.

\bibitem{salimans2022progressive}
Tim Salimans and Jonathan Ho.
\newblock Progressive distillation for fast sampling of diffusion models.
\newblock {\em arXiv preprint arXiv:2202.00512}, 2022.

\bibitem{teng2025magi}
Sand.ai.
\newblock Magi-1: Autoregressive video generation at scale.
\newblock {\em arXiv preprint arXiv:2505.13211}, 2025.

\bibitem{sarlin2020superglue}
Paul-Edouard Sarlin, Daniel DeTone, Tomasz Malisiewicz, and Andrew Rabinovich.
\newblock Superglue: Learning feature matching with graph neural networks.
\newblock In {\em IEEE Conf. Comput. Vis. Pattern Recog.}, 2020.

\bibitem{schonberger2016structure}
Johannes~L Schonberger and Jan-Michael Frahm.
\newblock Structure-from-motion revisited.
\newblock In {\em IEEE Conf. Comput. Vis. Pattern Recog.}, 2016.

\bibitem{gao2025seedance}
ByteDance Seed.
\newblock Seedance 1.0: Exploring the boundaries of video generation models.
\newblock {\em arXiv preprint arXiv:2506.09113}, 2025.

\bibitem{singer2022make}
Uriel Singer, Adam Polyak, Thomas Hayes, Xi~Yin, Jie An, Songyang Zhang, Qiyuan Hu, Harry Yang, Oron Ashual, Oran Gafni, Devi Parikh, Sonal Gupta, and Yaniv Taigman.
\newblock Make-a-video: Text-to-video generation without text-video data.
\newblock {\em arXiv preprint arXiv:2209.14792}, 2022.

\bibitem{song2023consistency}
Yang Song, Prafulla Dhariwal, Mark Chen, and Ilya Sutskever.
\newblock Consistency models.
\newblock {\em arXiv preprint arXiv:2303.01469}, 2023.

\bibitem{soomro2012ucf101}
Khurram Soomro, Amir~Roshan Zamir, and Mubarak Shah.
\newblock Ucf101: A dataset of 101 human actions classes from videos in the wild.
\newblock {\em arXiv preprint arXiv:1212.0402}, 2012.

\bibitem{transnetv2}
Tom{\'a}s Soucek and Jakub Lokoc.
\newblock Transnet v2: An effective deep network architecture for fast shot transition detection.
\newblock In {\em ACM Int. Conf. Multimedia}, 2024.

\bibitem{sun2025worldplay}
Wenqiang Sun, Haiyu Zhang, Haoyuan Wang, Junta Wu, Zehan Wang, Zhenwei Wang, Yunhong Wang, Jun Zhang, Tengfei Wang, and Chunchao Guo.
\newblock Worldplay: Towards long-term geometric consistency for real-time interactive world modeling.
\newblock {\em arXiv preprint arXiv:2512.14614}, 2025.

\bibitem{tang2025hunyuan}
Junshu Tang, Jiacheng Liu, Jiaqi Li, Longhuang Wu, Haoyu Yang, Penghao Zhao, Siruis Gong, Xiang Yuan, Shuai Shao, and Qinglin Lu.
\newblock Hunyuan-gamecraft-2: Instruction-following interactive game world model.
\newblock {\em arXiv preprint arXiv:2511.23429}, 2025.

\bibitem{team2023gemini}
Google~Gemini Team.
\newblock Gemini: a family of highly capable multimodal models.
\newblock {\em arXiv preprint arXiv:2312.11805}, 2023.

\bibitem{kong2024hunyuanvideo}
Hunyuan Foundation~Model Team.
\newblock Hunyuanvideo: A systematic framework for large video generative models.
\newblock {\em arXiv preprint arXiv:2412.03603}, 2024.

\bibitem{team2025longcat}
Meituan~LongCat Team.
\newblock Longcat-video technical report.
\newblock {\em arXiv preprint arXiv:2510.22200}, 2025.

\bibitem{mirage2}
Mirage Team.
\newblock Mirage 2.
\newblock \url{https://www.mirage2.org/}.
\newblock Accessed: 2026-01-26.

\bibitem{xiang2025pan}
PAN Team.
\newblock Pan: A world model for general, interactable, and long-horizon world simulation.
\newblock {\em arXiv preprint arXiv:2511.09057}, 2025.

\bibitem{bai2025qwen3}
Qwen Team.
\newblock Qwen3-vl technical report.
\newblock {\em arXiv preprint arXiv:2511.21631}, 2025.

\bibitem{chen2025seedance}
Seedance Team.
\newblock Seedance 1.5 pro: A native audio-visual joint generation foundation model.
\newblock {\em arXiv preprint arXiv:2512.13507}, 2025.

\bibitem{wan2025}
Wan Team.
\newblock Wan: Open and advanced large-scale video generative models.
\newblock {\em arXiv preprint arXiv:2503.20314}, 2025.

\bibitem{valevski2024diffusion}
Dani Valevski, Yaniv Leviathan, Moab Arar, and Shlomi Fruchter.
\newblock Diffusion models are real-time game engines.
\newblock {\em arXiv preprint arXiv:2408.14837}, 2024.

\bibitem{worldcanvas}
Hanlin Wang, Hao Ouyang, Qiuyu Wang, Yue Yu, Yihao Meng, Wen Wang, Ka~Leong Cheng, Shuailei Ma, Qingyan Bai, Yixuan Li, Cheng Chen, Yanhong Zeng, Xing Zhu, Yujun Shen, and Qifeng Chen.
\newblock The world is your canvas: Painting promptable events with reference images, trajectories, and text.
\newblock {\em arXiv preprint arXiv:2512.16924}, 2025.

\bibitem{spatialvid}
Jiahao Wang, Yufeng Yuan, Rujie Zheng, Youtian Lin, Jian Gao, Lin-Zhuo Chen, Yajie Bao, Yi~Zhang, Chang Zeng, Yanxi Zhou, Xiao-Xiao Long, Hao Zhu, Zhaoxiang Zhang, Xun Cao, and Yao Yao.
\newblock Spatialvid: A large-scale video dataset with spatial annotations.
\newblock {\em arXiv preprint arXiv:2509.09676}, 2025.

\bibitem{wang2025vggt}
Jianyuan Wang, Minghao Chen, Nikita Karaev, Andrea Vedaldi, Christian Rupprecht, and David Novotny.
\newblock Vggt: Visual geometry grounded transformer.
\newblock In {\em IEEE Conf. Comput. Vis. Pattern Recog.}, 2025.

\bibitem{wang2025koala}
Qiuheng Wang, Yukai Shi, Jiarong Ou, Rui Chen, Ke~Lin, Jiahao Wang, Boyuan Jiang, Haotian Yang, Mingwu Zheng, Xin Tao, Fei Yang, Pengfei Wan, and Di~Zhang.
\newblock Koala-36m: A large-scale video dataset improving consistency between fine-grained conditions and video content.
\newblock In {\em IEEE Conf. Comput. Vis. Pattern Recog.}, 2025.

\bibitem{xie2024physgaussian}
Tianyi Xie, Zeshun Zong, Yuxing Qiu, Xuan Li, Yutao Feng, Yin Yang, and Chenfanfu Jiang.
\newblock Physgaussian: Physics-integrated 3d gaussians for generative dynamics.
\newblock In {\em IEEE Conf. Comput. Vis. Pattern Recog.}, 2024.

\bibitem{xu2019understanding}
Jingjing Xu, Xu~Sun, Zhiyuan Zhang, Guangxiang Zhao, and Junyang Lin.
\newblock Understanding and improving layer normalization.
\newblock In {\em Adv. Neural Inform. Process. Syst.}, 2019.

\bibitem{yang2025longlive}
Shuai Yang, Wei Huang, Ruihang Chu, Yicheng Xiao, Yuyang Zhao, Xianbang Wang, Muyang Li, Enze Xie, Yingcong Chen, Yao Lu, Song Han, and Yukang Chen.
\newblock Longlive: Real-time interactive long video generation.
\newblock {\em arXiv preprint arXiv:2509.22622}, 2025.

\bibitem{yin2024improved}
Tianwei Yin, Micha{\"e}l Gharbi, Taesung Park, Richard Zhang, Eli Shechtman, Fredo Durand, and Bill Freeman.
\newblock Improved distribution matching distillation for fast image synthesis.
\newblock In {\em Adv. Neural Inform. Process. Syst.}, 2024.

\bibitem{yin2024one}
Tianwei Yin, Micha{\"e}l Gharbi, Richard Zhang, Eli Shechtman, Fredo Durand, William~T Freeman, and Taesung Park.
\newblock One-step diffusion with distribution matching distillation.
\newblock In {\em IEEE Conf. Comput. Vis. Pattern Recog.}, 2024.

\bibitem{yin2025slow}
Tianwei Yin, Qiang Zhang, Richard Zhang, William~T Freeman, Fredo Durand, Eli Shechtman, and Xun Huang.
\newblock From slow bidirectional to fast autoregressive video diffusion models.
\newblock In {\em IEEE Conf. Comput. Vis. Pattern Recog.}, 2025.

\bibitem{zhang2025matrix}
Yifan Zhang, Chunli Peng, Boyang Wang, Puyi Wang, Qingcheng Zhu, Fei Kang, Biao Jiang, Zedong Gao, Eric Li, Yang Liu, and Yahui Zhou.
\newblock Matrix-game: Interactive world foundation model.
\newblock {\em arXiv preprint arXiv:2506.18701}, 2025.

\bibitem{zhang2025waver}
Yifu Zhang, Hao Yang, Yuqi Zhang, Yifei Hu, Fengda Zhu, Chuang Lin, Xiaofeng Mei, Yi~Jiang, Bingyue Peng, and Zehuan Yuan.
\newblock Waver: Wave your way to lifelike video generation.
\newblock {\em arXiv preprint arXiv:2508.15761}, 2025.

\bibitem{zhao2023pytorch}
Yanli Zhao, Andrew Gu, Rohan Varma, Liang Luo, Chien-Chin Huang, Min Xu, Less Wright, Hamid Shojanazeri, Myle Ott, Sam Shleifer, Alban Desmaison, Can Balioglu, Pritam Damania, Bernard Nguyen, Geeta Chauhan, Yuchen Hao, Ajit Mathews, and Shen Li.
\newblock Pytorch fsdp: experiences on scaling fully sharded data parallel.
\newblock {\em arXiv preprint arXiv:2304.11277}, 2023.

\bibitem{zitkovich2023rt}
Brianna Zitkovich, Tianhe Yu, Sichun Xu, Peng Xu, Ted Xiao, Fei Xia, Jialin Wu, Paul Wohlhart, Stefan Welker, Ayzaan Wahid, et~al.
\newblock Rt-2: Vision-language-action models transfer web knowledge to robotic control.
\newblock In {\em Conf. on Robot Learn.}, 2023.

\end{thebibliography}
}

\end{document}